\documentclass[journal]{IEEEtran}

\usepackage[T1]{fontenc}

\ifCLASSINFOpdf
\else
\fi

\usepackage{cite}
\usepackage{amsmath}
\usepackage{color}
\usepackage[cmintegrals]{newtxmath}

\usepackage{tabularx}
\usepackage{multirow}
\usepackage{booktabs}
\usepackage{graphicx}
\usepackage{algorithm}
\usepackage{algorithmic}
\usepackage{color}
\usepackage{hyperref}

\usepackage{amsthm}
\theoremstyle{definition}

\newtheorem{proof of theorem}{Proof of Theorem}
\newtheorem{definition}{Definition}
\usepackage{threeparttable}
\usepackage{subfigure}
\usepackage{caption}
\usepackage{color}

\begin{document}

\title{Effective Tensor Completion via Element-wise Weighted Low-rank Tensor Train with Overlapping Ket Augmentation}

\author{Yang Zhang, Yao~Wang,~\IEEEmembership{Member,~IEEE,}
                  Zhi~Han,~\IEEEmembership{Member,~IEEE,} Xi'ai~Chen,~Yandong~Tang 
\thanks{Yang Zhang,  Zhi Han, Xi'ai Chen and Yandong Tang are with the State Key Laboratory of Robotics, Shenyang Institute of Automation, Chinese Academy of Sciences, Shenyang 110016, P.R. China. (email: hanzhi@sia.cn)}
\thanks{Yao Wang is with the Center for Intelligent Decision-making
and Machine Learning, School of Mangement, Xi’an Jiaotong University, Xi’an 710049, P.R. China. (email: yao.s.wang@gmail.com)}}
\maketitle

\begin{abstract}
In recent years, there have been an increasing number of applications of tensor completion based on the tensor train (TT) format because of its efficiency and effectiveness in dealing with higher-order tensor data. However, existing tensor completion methods using TT decomposition have two obvious drawbacks. One is that they only consider mode weights according to the degree of mode balance, even though some elements are recovered better in an unbalanced mode. The other is that serious blocking artifacts appear when the missing element rate is relatively large. To remedy such two issues, in this work, we propose a novel  tensor completion approach via the element-wise weighted technique. Accordingly, a novel  formulation for tensor completion and an effective optimization algorithm, called as tensor completion by parallel weighted matrix factorization via tensor train (TWMac-TT),  is proposed. In addition, we specifically consider the recovery quality of edge elements from adjacent blocks. Different from traditional reshaping and ket augmentation, we utilize a new tensor augmentation technique called overlapping ket augmentation, which can further avoid blocking artifacts. We then conduct extensive performance evaluations on synthetic data and several real image data sets. Our experimental results demonstrate that the proposed algorithm TWMac-TT outperforms several other competing tensor completion methods. The code is available at \url{https://github.com/yzcv/TWMac-TT-OKA}.
\end{abstract}


\IEEEpeerreviewmaketitle

\section{Introduction}
\IEEEPARstart{T}{ensors} are higher-order generalizations of matrices and vectors, which are represented as multidimensional arrays. Thus, tensors possess a better ability to represent practical multidimensional data, {such as} RGB images, hyperspectral images and video sequences, compared with matrices and vectors. Generally, although such tensors are residing in extremely high-dimensional spaces, they often have low-dimensional structures that can naturally be characterized by low-rankness. Consequently, low-rank tensor modeling is a powerful technique in practical multidimensional data analysis and has received much attention in recent years, e.g., \cite{vasilescu2003multilinear,sun2005cubesvd,franz2009triplerank}.

As a generalization of low-rank matrix completion (LRMC) \cite{cai2010singular,ma2011fixed,recht2010guaranteed}, low-rank tensor completion (LRTC) aims at recovering the missing entities of a higher-order tensor whose entries are partially observed\cite{bertalmio2000image,komodakis2006image,korah2007spatiotemporal}. It has achieved great success in the fields of computer vision, signal processing, and machine learning, among numerous others~\cite{liu2013tensor,signoretto2010nuclear,signoretto2011tensor,gandy2011tensor,tomioka2011statistical,tan2014tensor,xu2013parallel,xu2020factorized,yuan2018high,ko2020fast}. Due to the ability to maintain the intrinsic structures  of the data,  LRTC methods generally {outperform} LRMC methods with natural multidimensional image and video data, e.g.,~\cite{liu2013tensor,zhang2016exact,7502115,bengua2017efficient,zhou2017tensor,zhang2019nonlocal,chang2020weighted}. In addition, the most popular LRTC methods are based on CANDECOMP/PARAFAC (CP) decomposition \cite{carroll1970analysis,harshman1970foundations}, and Tucker decomposition~\cite{tucker1966some}. However, for CP, it is still hard to compute the CP rank due to the NP-hard nature. The difficulties of CP rank estimation impede the wider applications of CP decomposition to a large extent. For Tucker, by unfolding an $N$-way tensor into $N$ matrices, the obtained matrices are comparably unbalanced. In other words, there is a big gap between the number of rows and columns of the unfolding matrices. This nature of imbalance hidden in Tucker results in the inefficiency of Tucker rank minimizations, which further limits its performance in dealing with the tensor completion task in practice.

As recently stated in~\cite{bengua2017efficient}, tensor train (TT) decomposition \cite{oseledets2011tensor} could overcome the aforementioned shortcomings of CP and Tucker decompositions. Specifically, the seminal work \cite{bengua2017efficient} demonstrated that tensor completion algorithms based on TT decomposition perform better than other popular algorithms in dealing with several image processing tasks. Then TT decomposition has been widely applied in some computer vision tasks {\cite{bengua2017efficient,bengua2016concatenated,TNN_tt_superresolution,TNN_tt_analysis,TNN_tt_regression, yuan2018high, ko2020fast}}. However, such popular TT decomposition algorithms need to address two critical issues:
\begin{enumerate}
\item Weights assignment. The component matrices obtained by TT decomposition include both balanced and unbalanced matrices. Among them, the unbalanced matrices refer to  the matrices that are uneven in the number of rows and columns. As stated in~\cite{bengua2017efficient}, balanced matrices generally perform better at dealing with tensor recovery tasks. Hence, balanced component matrices are given larger weights when folding matrices back to tensors. However, according to our observation, even in the most unbalanced mode, some elements are recovered more precisely than those in the balanced mode, which will be illustrated in detail in Section \ref{model}. To sum up, the current weights assignment strategy is rough and inaccurate to some extent.
\item Tensor augmentation. 
It is necessary for low-TT-rank approximation to augment the order of the tensor. As the order of tensor increases, the number of factor matrices increases, and more relatively balanced matrices can be obtained for further low-rank recovery. In this way, TT-rank minimization becomes more effective as we try to optimize the objective from multiple perspectives, i.e., multiple sub-matrices from the original tensor. For instance, compared with Tucker decomposition, given a three-order tensor data, the number of TT ranks degrades to two, which is a subset of the Tucker rank (which is three). Thus, TT decomposition does not perform better than Tucker decomposition in practice. 
It is vital to make the tensor order higher than three so that we can make better use of TT decomposition. This necessitates some tensor augmentation schemes, including vanilla reshaping \cite{wang2017efficient} and the ket augmentation (KA) \cite{bengua2017efficient,bengua2016concatenated,latorre2005image}. Though KA possesses better physical meaning~\cite{latorre2005image} than the reshaping technique, in the order-increasing procedure of KA, the lower-order tensor is evenly divided into several blocks without utilizing any neighborhood information. Therefore, once the missing element rate is high, the recovered tensors augmented by KA often have apparent blocking artifacts,  just as our experimental study shows in Section \ref{exp}. 

\end{enumerate}

To overcome the aforementioned drawbacks, we focus on both issues. Firstly, the current mode-wise weights assignment scheme cannot properly use the recovered information of each mode, as the recovery quality of every element does not strictly accord with the degree of balance of the mode matrix to which it belongs. To encourage well-recovered elements in the unbalanced mode and suppress poorly recovered elements in the balanced mode, we consider the weights of elements rather than the weights of modes. 

Secondly, to allow tensor augmentation to maintain neighborhood information better, we propose a new augmentation scheme by introducing the overlapping idea. The benefits of such {a} new scheme are two-fold: 1) the order of the tensor can be further increased compared with that by traditional methods, which plays a vital role in low-TT-rank optimization; 
and 2) the overlapping procedure enforces local compatibility and smoothness constraints\cite{chang2004super}, which guarantees that the use of neighborhood information can avoid the blocking effect.

The main contributions of this work can be summarized as follows:
\begin{itemize}
\item We propose an element-wise weighted low-rank tensor completion via tensor train (EWLRTC-TT) model for dealing with the LRTC task, in which each element can be estimated more precisely. Additionally, to prevent over-smoothing of the overlapping regions, the elements in the overlapping regions in each component matrix are assigned different weights.

\item We derive an effective algorithm for solving the EWLRTC-TT model called tensor completion by weighted parallel matrix factorization via tensor train (TWMac-TT). By parallel computing via multiple workers, the algorithm can guarantee high computation efficiency. 

\item We propose a novel tensor augmentation scheme called overlapping ket augmentation (OKA) and then incorporate it into the proposed EWLRTC-TT model. The experimental results demonstrate that the EWLRTC-TT model combined with OKA significantly outperforms several state-of-the-art methods.

\end{itemize}

The rest of the paper is organized as follows. In Section \ref{notation},  we introduce the notations and tensor basics used throughout the paper.
The formulations of the EWLRTC-TT model and the OKA scheme are proposed in Section \ref{framework}. In addition, an effective solving algorithm called TWMac-TT is also presented. In Section \ref{exp}, we present extensive experimental results to show how our proposed model outperforms other popular alternatives. The conclusion of our work is finally summarized in Section \ref{conclusion}.

\section{Notations and basic definitions}
\label{notation}

\begin{definition}[tensor]
Tensor\cite{kolda2009tensor} is a high-order generalization of vector and matrix, whose dimension is called \textit{order} or \textit{mode}. It can be understood as a multi-dimensional array. In this paper, scalers, vectors, and matrices are denoted by lowercase letters $(a,b,c,\dots)$, boldface lowercase letters $(\mathbf{a,b,c,}\dots)$, and capital letters $(A,B,C,\dots)$, respectively. Higher-order tensors which means their orders are three or above are denoted by calligraphic letters $(\mathcal{A,B,C,}\dots)$.
\end{definition}

\begin{definition}[Frobenius norm]
 An $N$th-order tensor is represented as $\mathcal{X} \in \mathbb{R}^{I_1\times I_2 \times \cdots \times I_N}$ with elements $x_{i_1\cdots i_k\cdots i_N}$, where $I_k (k=1,\dots,N)$ is the dimension corresponding to mode $k$. The Frobenius\cite{kolda2009tensor} norm of $\mathcal{X}$ is defined as
\begin{equation}
||\mathcal X||_F = \sqrt{\sum_{i_1}\sum_{i_2}\cdots\sum_{i_N}x_{i_1i_2\cdots i_N}^2}.
\end{equation}
\end{definition}

\begin{definition}[mode-$k$ canonical matricization]
{Mode-$k$ canonical matricization \cite{cichocki2014tensor} of a tensor $\mathcal{X} \in \mathbb{R}^{I_1\times I_2 \times \cdots \times I_N}$, denoted as $\mathcal X_{<k>}$, is an operation that reshapes the tensor into a matrix by putting the first $k$ modes in the matrix rows and {putting} the remaining modes in the columns, i.e.,
\begin{equation}
\label{canonical}
\mathcal X_{<k>} \in \mathbb R^{(\prod_{t=1}^k I_t)\times (\prod_{t=k+1}^N I_t)},
\end{equation}
such that 
\begin{align}
&\mathcal X(i_1,\cdots,i_N) \notag \\
&\quad\quad\quad= \mathcal X_{<k>}(i_1+(i_2 -1)I_1+\cdots+(i_k-1)\prod_{t=1}^{k-1}I_t),\notag\\
&\quad\quad\quad i_{k+1}+(i_{k+2}-1)I_{k+1}+\cdots+(i_N-1)\prod_{t=k+1}^{N-1}I_t).
\end{align}}

{To be noticed, we use \textbf{the mode matrix or the mode} to refer to the mode-$k$ canonical matrix $\mathcal{X}_{<k>}$ in the rest of the paper.}
\end{definition}

\begin{definition}[tensor train]\label{tt_formulation}
Let $ \mathcal{X}\in \mathbb{R}^{I_1\times I_2\times \cdots \times I_N}$ be an \textit{N}-th order tensor with $I_i$ dimension along the $i$-th mode, each element of the tensor can be represented in the form of tensor train (TT)\cite{oseledets2011tensor}:
\begin{equation}
\begin{split}
&\mathcal X(i_1,\cdots,i_n) = \\
&\sum_{r_0=1}^{R_0} \cdots \sum_{r_N=1}^{R_N} G_1(r_0,i_1,r_1)G_2(r_1,i_2,r_2)\cdots G_N(r_{N-1},i_N,r_N),
\end{split}
\label{tt_eq}
\end{equation}
where $G_i\in \mathbb{R}^{R_{i-1}\times I_i \times R_i}$ is a set of 3-order tensors. And  the vector $[R_0,\cdots,R_N]$ is the tensor train rank (TT-rank).
\end{definition}

\section{The proposed framework} 
\label{framework}
\subsection{EWLRTC-TT Model}
\label{model}
The goal of matrix completion is to recover missing entries of a matrix $T\in\mathbb R^{m\times n}$ from its partially known entries given by a subset $\Omega$. This can be achieved via the well-known rank minimization technique\cite{fazel2002matrix,kurucz2007methods}:
\begin{equation}
\label{LRMC}
\min_X \text{rank}(X)\quad s.t.\quad X_{\Omega} = T_{\Omega}.
\end{equation}

{Eq. \eqref{LRMC} formulates the low rank matrix completion (LRMC) problem.} As a higher-order generalized form of matrix completion, the optimization problem of tensor completion can be similarly written as follows:
\begin{equation}\label{LRTC}
\min_{\mathcal X} \text{rank}(\mathcal X)\quad s.t.\quad \mathcal X_{\Omega}=\mathcal T_{\Omega},
\end{equation}
where $\mathcal T \in\mathbb R^{I_1\times I_2\times \cdots \times I_N}$ is an $N$th-order tensor representing the true tensor, and the index set $\Omega$ gives the location of partial known entries. {Eq. \eqref{LRTC} is called the low rank tensor completion (LRTC) problem.}

Based on tensor train decomposition and parallel matrix factorization, the optimization objective of the tensor completion problem can be written as
{
\begin{equation}
\min_{U_k,V_k,\mathcal X} \sum_{k=1}^{N-1} \frac{\alpha_k}{2}||U_kV_k-\mathcal{X}_{<k>}||_F^2\quad s.t.\quad \mathcal X_{\Omega}=\mathcal T_{\Omega},
\label{objectfunction}
\end{equation}
}which is called TMac-TT in \cite{bengua2017efficient}. In this model, {the mode matrices $\{\mathcal{X}_{<k>}, k=1,\cdots,N-1\}$ are obtained by Eq. \eqref{canonical}. For simplicity, let $m_k = \prod_{l=1}^k I_l, n_k = \prod_{l=k+1}^N I_l$ representing the number of rows and columns of $\mathcal{X}_{<k>}$ respectively.} {$\mathcal{X}_{<k>} = U_k V_k$ denotes the parallel matrix factorization for mode matrix $\mathcal{X}_{<k>}$, where $U_k\in\mathbb R^{m_k\times r_k}$, $V_k\in \mathbb R^{r_k\times n_k}$.} And $\alpha_k$ denotes the weight of the matrix $\mathcal{X}_{<k>}$ with $\sum_{k=1}^{N-1} \alpha_k =1$. The value of the weight $\alpha_k$ is calculated according to the degree of balance of the corresponding {mode matrix $\mathcal{X}_{<k>}$}. { In other words, we assign larger weights to the more balanced factor matrices and smaller weights to the less balanced factor matrices.}
The balance degree {for mode-$k$ canonical factor matrix $\mathcal{X}_{<k>}$} is defined by
{
\begin{equation}
  \alpha_{k}=\frac{\xi_{k}}{\sum_{k=1}^{N-1} \xi_{k}}, \quad \xi_{k}=\min \left(m_k, n_k\right),
\end{equation}}
where $k=1, \ldots, N-1$, { and $m_k = \prod_{l=1}^k I_l, n_k = \prod_{l=k+1}^N I_l$ as we defined earlier.} It has been shown that this TMac-TT model outperforms other { TT-based tensor completion models, namely SiLRTC-TT, SiLRTC-Square and TMac-Square}~\cite{bengua2017efficient}.

Even though the most balanced mode is assigned the most prominent weight based on the assumption that it performs the best among all the modes, unbalanced modes are also considered to contribute to the final recovery result. {In practice, the recovery quality obtained by integrating the low-rank approximation of all the factor modes is significantly better than that obtained by using only the most balanced mode. } This fact implies that unbalanced modes also provide essential information. We believe the accuracy of recovery of some elements in unbalanced modes is even better than that obtained in the most balanced mode.

\begin{figure}[ht]
\centering
\subfigure{\includegraphics[width=0.2\textwidth]{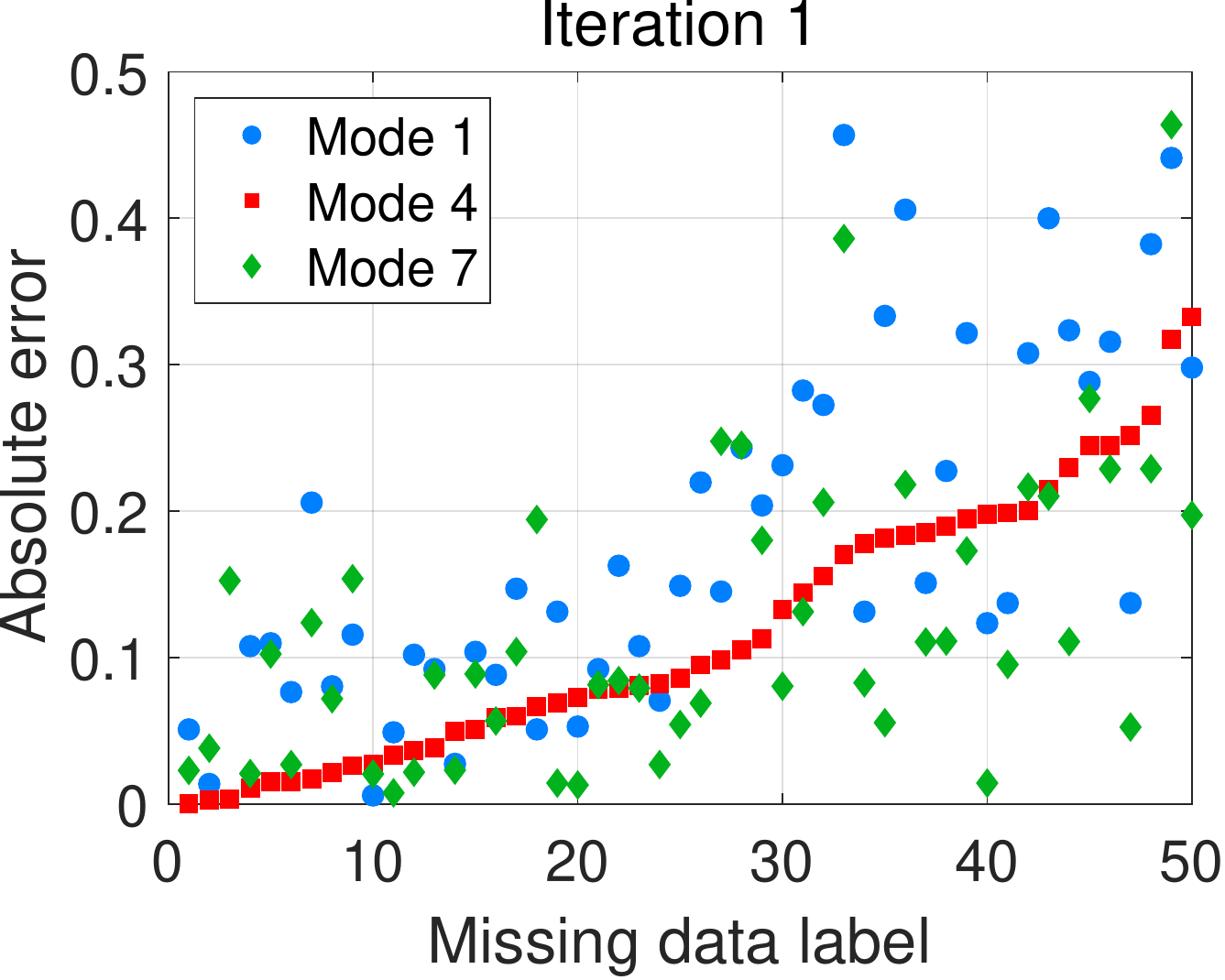}}
\subfigure{\includegraphics[width=0.2\textwidth]{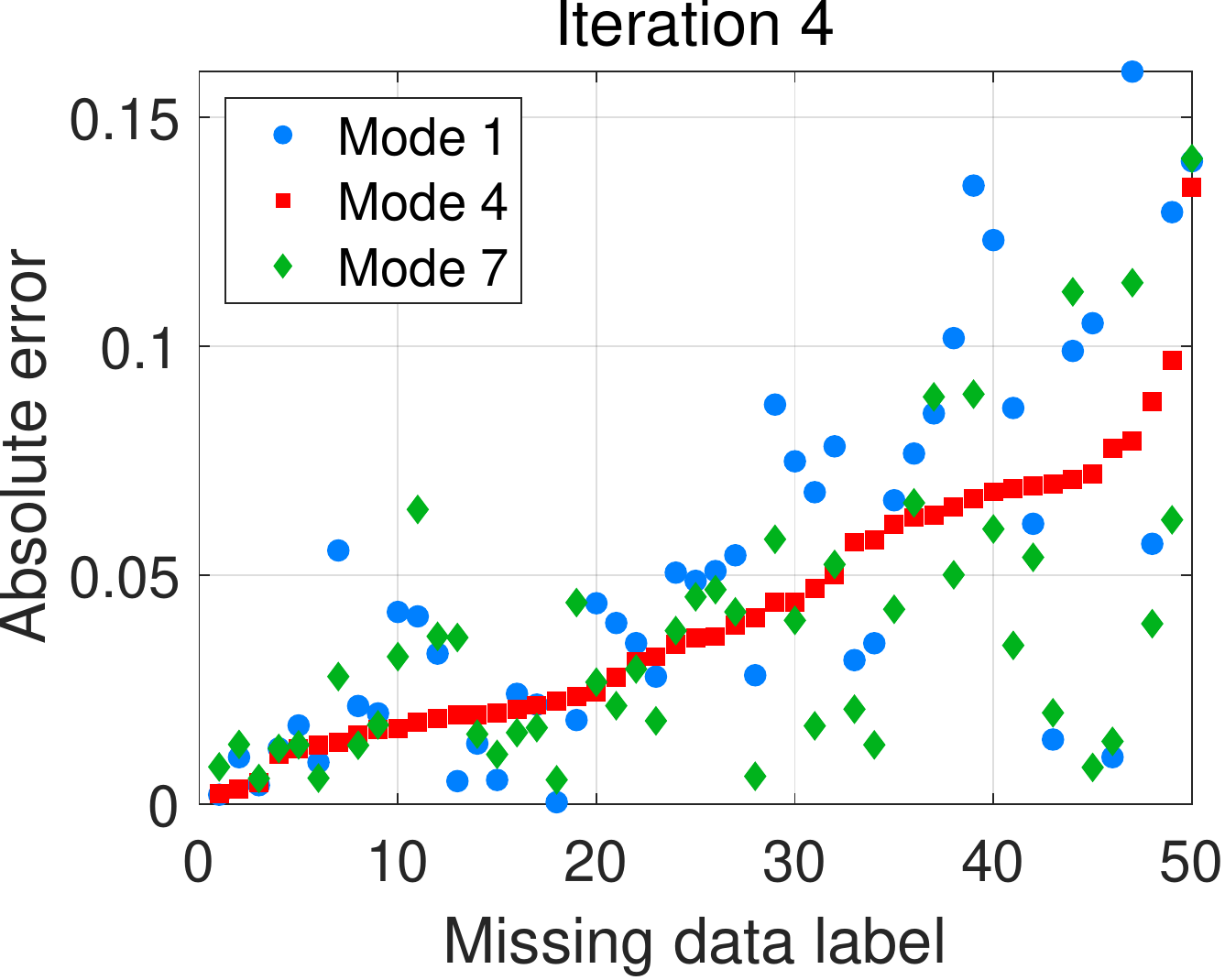}}
\subfigure{\includegraphics[width=0.2\textwidth]{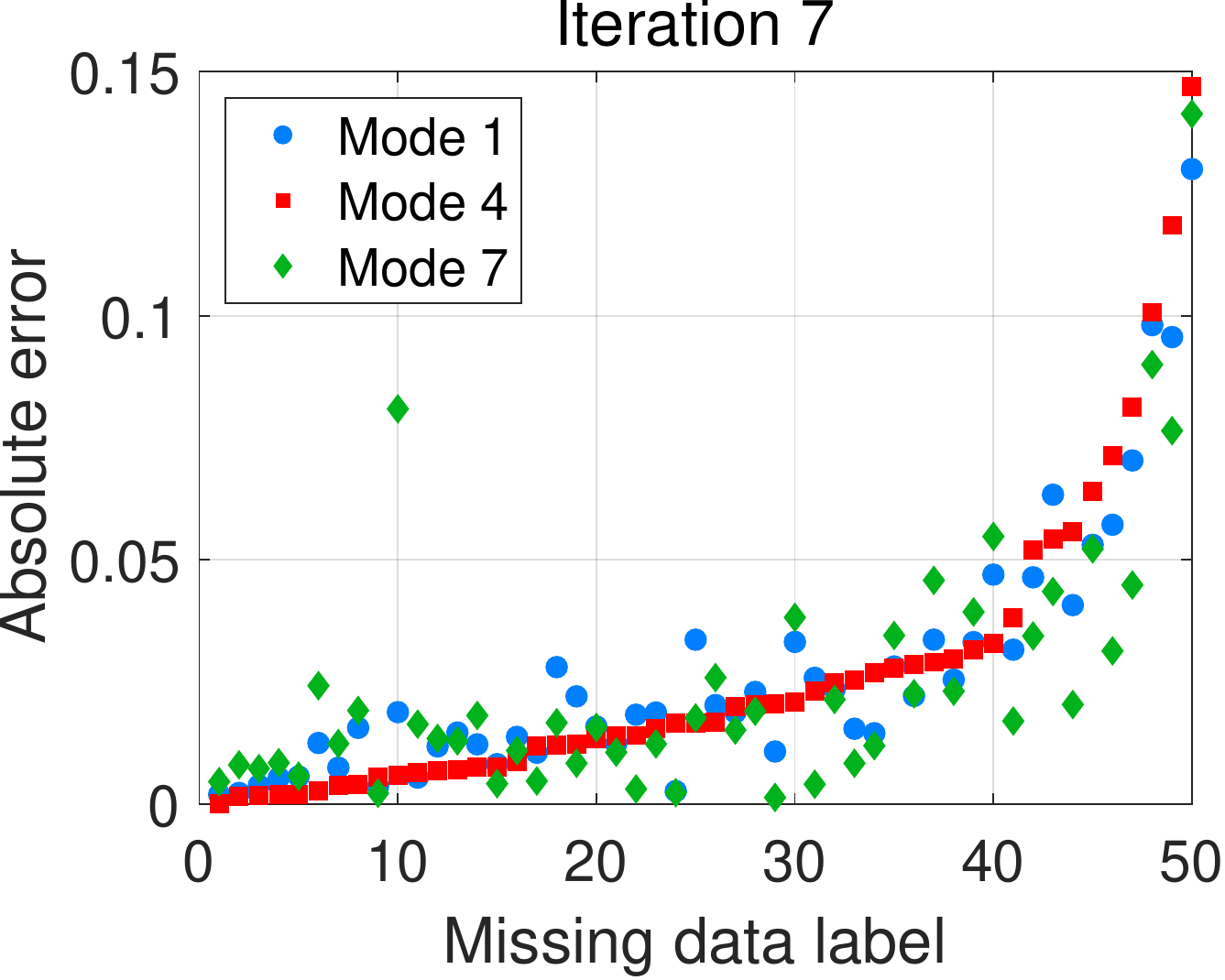}}
\subfigure{\includegraphics[width=0.2\textwidth]{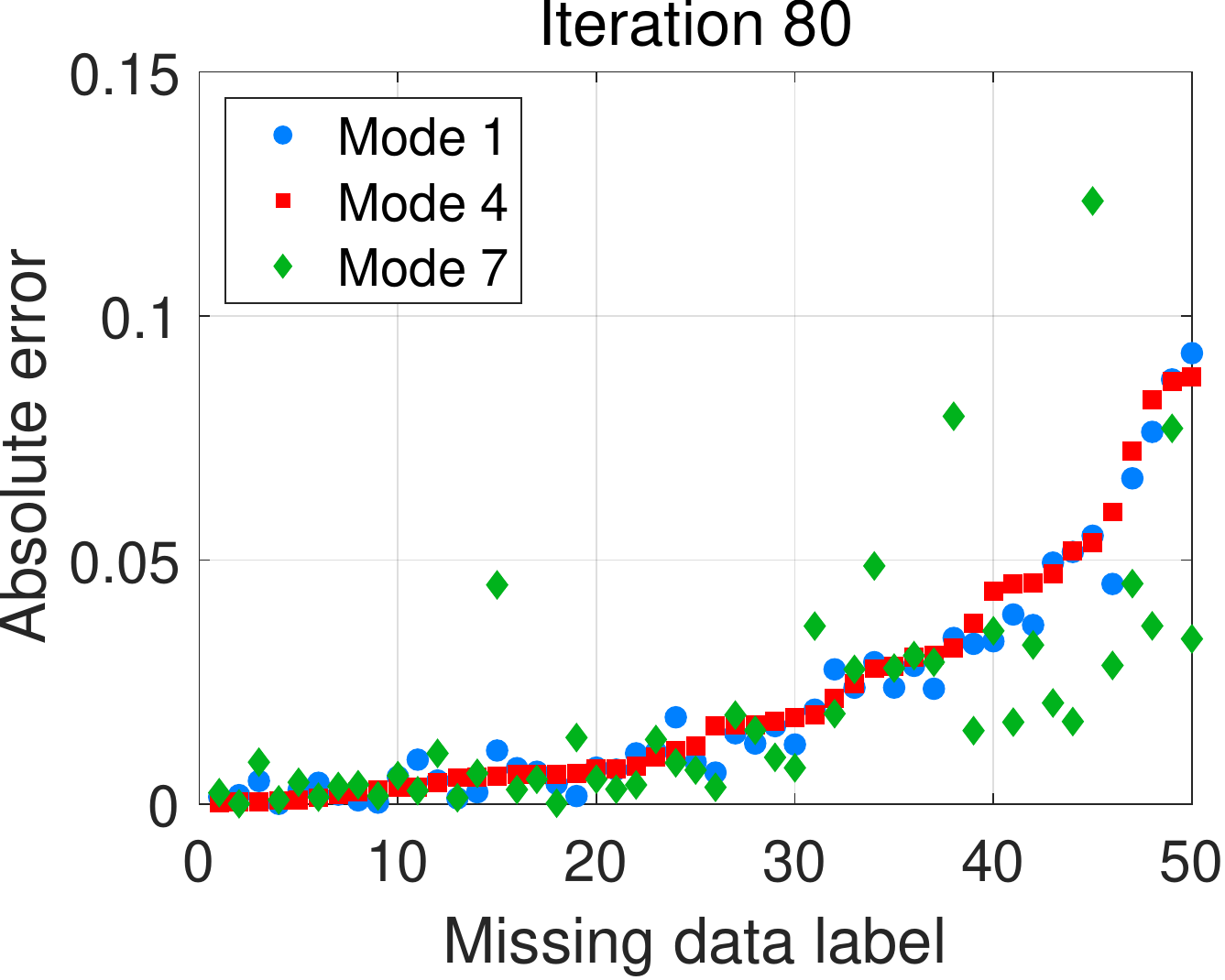}}
\caption{
{The absolute error $\{\delta_k, k=1,4,7\}$ of missing elements recovered in the mode-k matrix. Matrix size (1) Mode 1 (unbalanced): $4\times4^7$; (2) Mode 4 (balanced): $4^4\times4^4$; (3) Mode 7 (unbalanced): $4^7\times4$.}}
\label{ModevsElement}
\end{figure}

To further demonstrate this conjecture, {we use the absolute error to assess the quality of the recovery for each mode matrix. Specifically, we construct an incomplete tensor $\mathcal T$ and the corresponding location indicator $\Omega$ by randomly sampling the ground-truth $\mathcal{T}_{\text{true}}$. Then we use TMac-TT to estimate $\mathcal{X}_{<k>}$ iteratively by optimizing Eq. \eqref{objectfunction}. During each iteration, we can calculate the absolute error  $\delta_k = \left|\mathcal{T}_{\text{true}<k>}-\mathcal{X}_{<k>}\right|$, where  $\delta_k$ denotes the absolute error matrix for mode $k$. By observing the absolute errors of a missing element recovered from different mode matrices, we can recognize the actual completion performance in each mode matrix for a missing element.}

We take a real gray image Lena of size $256\times 256$ as an example, to figure out whether the most balanced mode can best restore all the missing elements. To better use the ability of TT, before the image completion process, we use the KA to increase the order of the image and get an eighth-order tensor $\mathcal T_{\text{true}} \in \mathbb R^{4\time 4\times 4\times 4\times 4\times 4\times 4\times 4\times 4}$. {We randomly sampled 50\% missing elements as known and get $\mathcal{T}$ and $\Omega$.} 
{By canonically matricizing the tensor $\mathcal T_{\text{true}}$ and  $\mathcal T$ using Eq.~\eqref{canonical}, mode matrices  $\{\mathcal{T}_{\text{true}<k>},~k=1,\cdots, 7\}$ and $\{\mathcal{T}_{<k>},~k=1,\cdots, 7\}$ are obtained, respectively}.  
{Among all the missing elements, we randomly choose 50 elements for illustration in Fig. \ref{ModevsElement}. } For better visualization, we {sort the absolute errors of the missing elements recovered from the most balanced mode (i.e., elements in $\delta_4$ for mode 4 {matrix $\mathcal{X}_{<4>}$ of size $4^4\times 4^4$}) in ascending order, and plot the absolute errors $\delta_k$ of the corresponding elements estimated from the unbalanced modes (i.e., elements in $\delta_1$ and $\delta_7$ for modes 1 and 7)}.

From Fig. \ref{ModevsElement}, we observe that, although mode 4 performs best among all the modes for most of the elements, there is still some percentage of the elements in the unbalanced modes that provide better recovery. {We choose a few iterations at the early stage (iteration 1, 4, and 7) of the TMac-TT algorithm and an iteration (iteration 80) before the algorithm converges to verify our conjecture. The phenomenon mentioned above will not disappear as the algorithm iterates.} { Thus, we conclude that it is not true that the most balanced mode can produce the best recovery for all the elements no matter how many iterations are.} On the other hand, we can observe that the absolute error from unbalanced modes has a large variance at first a few iterations and tends to converge as the algorithm iterates more. This phenomenon is in line with our common sense that the balanced mode initially provides a stronger rank constraint. Yet, the unbalanced modes tend to reveal more information more stably when most elements have been appropriately recovered.
{To sum up, the mode weight assignment scheme is inaccurate and may degrade the restoration quality.}

Therefore, we consider estimating weight for each element. In this way, accurate element-wise weights can be calculated iteratively by the proposed model.

{The proposed} approach to the LRTC problem {in Eq. \eqref{LRTC}} is {the so-called EWLRTC-TT model.} Firstly, we formulate the LRTC-TT model, which is closely to the proposed EWLRTC-TT model. The matrix decomposition for the $k$-th mode can be represented as $\mathcal{X}_{<k>} = U_k {V_k}'$ for $\mathcal{X}_{<k>}\in\mathbb R^{m_k\times n_k}$, where $U_k\in\mathbb R^{m_k\times r_k}$ and $V_k\in \mathbb R^{n_k\times r_k}$. Any rank-$r$ matrix can be decomposed in such a way and any pair of such smaller matrices could yield a rank-$r$ matrix by multiplying two such factors. {Therefore, the LRTC-TT can be disassembled into sub-problems over pairs of matrices $(U_k, V_k)$ in the $k$-th mode with the following optimization objective:}
{
\begin{equation}
\begin{aligned}
\min_{U_k,V_k,\mathcal{X}_{<k>}} &||M_k\odot (U_k {V_k}' - \mathcal{X}_{<k>})||_F^2,\\ 
&s.t.\quad (\mathcal{X}_{<k>})_{\Omega} = (\mathcal{T}_{<k>})_{\Omega},\\
\end{aligned}
\end{equation}
}where $\mathcal{X}_{<k>} = [x_{ij}] \in \mathbb R^{m_k\times n_k}$ is the given {incomplete matrix obtained from a high-order incomplete tensor by TT decomposition;} {$ M_k=[m_{ij}] \in \mathbb R^{m_k\times n_k}$ is a binary matrix for the incomplete $\mathcal{X}_{<k>}$, where $m_{ij} = 1$ if $x_{ij}$ is known and $m_{ij}=0$ if $x_{ij}$ is missing; $\odot$ represents the Hadamard product.} 
{Via combining the different mode matrices together, the overall objective function is obtained, that is, }
{
\begin{equation}
\min_{U_k,V_k,\mathcal{X}} \sum_{k=1}^{N-1}||M_k\odot (U_k {V_k}' - \mathcal{X}_{<k>})||_F^2, \quad s.t.\quad \mathcal{X}_\Omega = \mathcal{T}_\Omega.
\end{equation}
}

{Secondly, the EWLRTC-TT model is formulated as follows. Different from the fixed indicator matrix $M_k$ in the $k$-th mode, EWLRTC-TT adopts the automatically updated weight $W_k$ for the $k$-th mode matrix and the objective can thus be formulated as}
{
\begin{equation}
\begin{aligned}
\min_{U_k,V_k,\mathcal X, W_k}  J = &\sum_{k=1}^{N-1}||W_k\odot (U_kV_k-X_{<k>})||_F^2,\\
&s.t.\quad \mathcal{X}_\Omega = \mathcal{T}_\Omega,
\label{objectfunctionours}
\end{aligned}
\end{equation}
}{where $W_k$ no longer remains static with zeros and ones but is filled with estimated $w_{ij} \in (0, 1)$ for the missing locations and $w_{ij} = 1$ if the element $x_{ij}$ is known.}

\subsection{Overlapping ket augmentation}

\begin{figure}[htp]
\centering
\includegraphics[width=3.2in]
{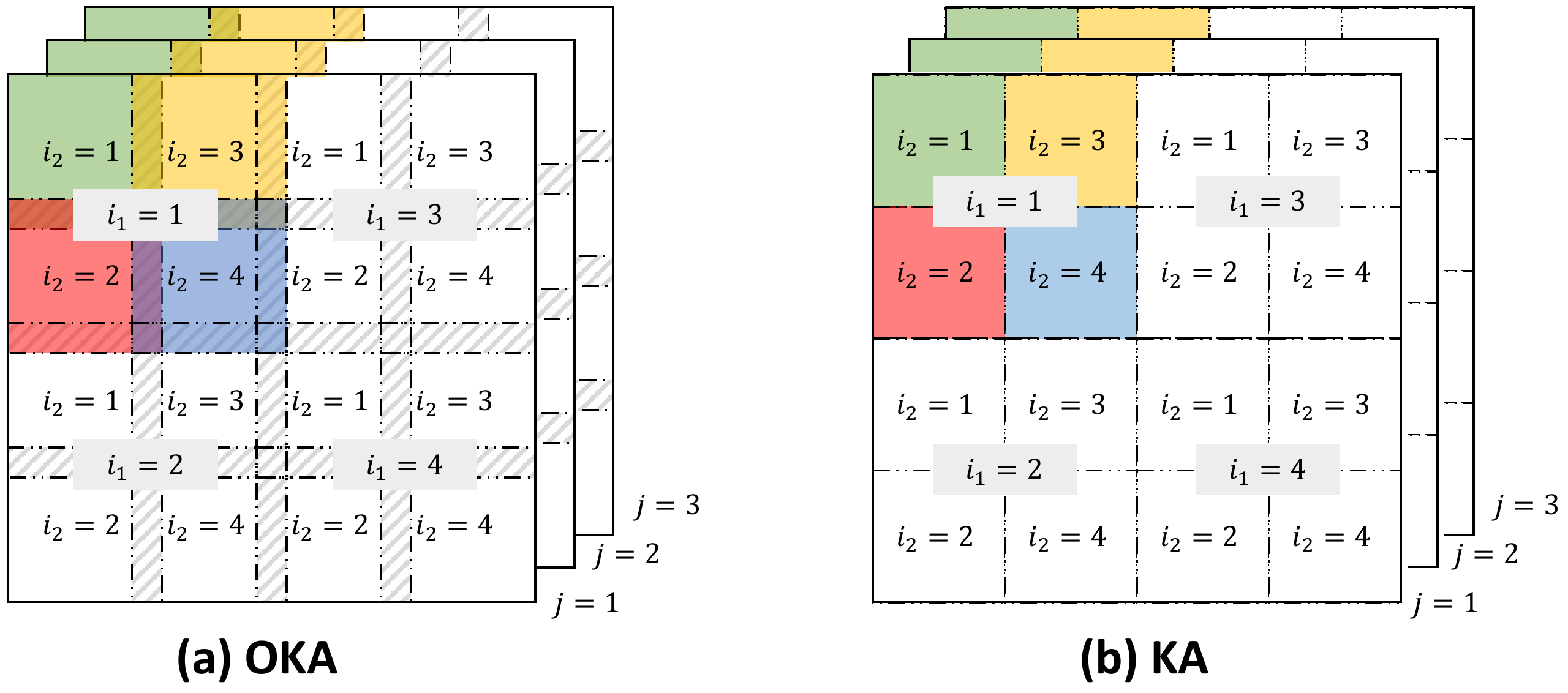}
\caption{Process of OKA and KA: example for an RGB image of third-order into a fifth-order tensor.}
\label{OKA}
\end{figure}

To eliminate the blocking artifacts caused by the KA, a new manner of tensor augmentation called overlapping ket augmentation (OKA) is proposed. Fig. \ref{OKA} shows the comparison between OKA and KA using an RGB image. Specifically, we assume that the color is indexed by $j$, where $j = 1, 2,3$ represents the channels of an RGB image. Different from KA, OKA divides the matrices from three channels into four blocks with elements overlapped. The sub-blocks obtained by the first division are indexed by $i_1$. Next, taking one sub-block marked with colors as an example, OKA further divides the colored block into four smaller overlapped sub-blocks marked with different colors and retrieves them by index $i_2$. A higher-order tensor can be constructed by repeating such division steps.

\begin{figure*}[htb]
\centering
\includegraphics[width=.65\textwidth]{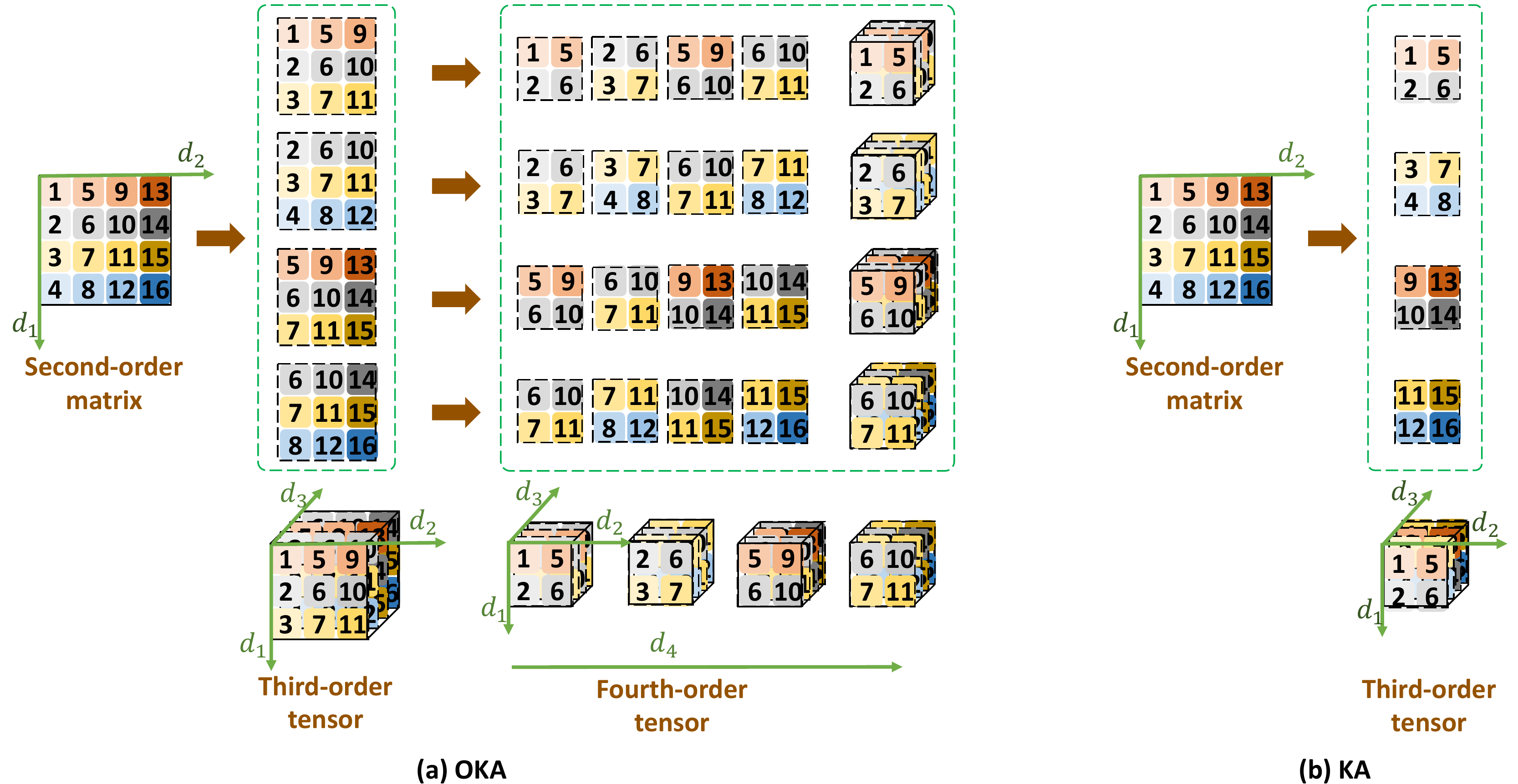}
\caption{{The structured block addressing procedure to cast a second-order matrix of size 4$\times$4 into a higher-order tensor. (a) Tensor augmentation by OKA yields a tensor of size $4\times 4\times 4 \times 4$. There are two steps to increase the tensor order. By the first step, a third-order tensor is obtained with three dimension marked with $d_1$, $d_2$ and $d_3$. By the second step, a fourth-order tensor is formed from the third-order tensor with four dimension denoted with $d_1$, $d_2$, $d_3$ and $d_4$. (b) Tensor augmentation by KA yields a tensor of size $4\times 4\times 4$. Without any element overlapped, only one step is taken to arrive at a third-order tensor by KA.}}
\label{OKAKA}
\end{figure*}

Fig. \ref{OKAKA} further illustrates the element reallocation procedure of OKA and KA when we transform a matrix $\mathcal{T} \in \mathbb R^{4\times4}$ into a higher-order tensor. The small squares marked with colors and numbers in Fig. \ref{OKAKA} represent different elements of the matrix. Fig.~\ref{OKAKA}(a) shows the procedure of OKA. The matrix is divided with two overlapped elements to get four sub-matrices, which can be stacked into a third-order tensor of size $3\times 3\times 4 $ in the first step. Then, OKA further divides this third-order tensor with one overlapped element to form four third-order tensors, and stacks them into a fourth-order tensor of size $4\times 4\times 4\times 4$. KA applies a similar procedure except no element overlapped when dividing the matrix. Thus, we get a third-order tensor by KA as shown in Fig.~\ref{OKAKA}(b).

The process in Fig. \ref{OKAKA}(a) can be formulated mathematically as follows,
\begin{equation}
     \mathcal{T}_{[2^2\times 2^2]} = \sum_{i_4=1}^4\sum_{i_3=1}^4\sum_{i_2=1}^4\sum_{i_1=1}^4 c_{i_4 i_3 i_2 i_1} \mathbf{e}_{i_4}\otimes \mathbf{e}_{i_3}\otimes \mathbf{e}_{i_2}\otimes \mathbf{e}_{i_1},
\end{equation}
where $c_{i_4 i_3 i_2 i_1}$ is the pixel value in the $4\times 4$ matrix indexed by the process shown in Fig.~\ref{OKAKA}(a), and $\mathbf{e}_{i_k} (k=1, 2, 3, 4)$ is the orthonormal base, which has the same meaning as that in the KA formulation.

\begin{figure*}[htb]
\centering
\includegraphics[width=6in]{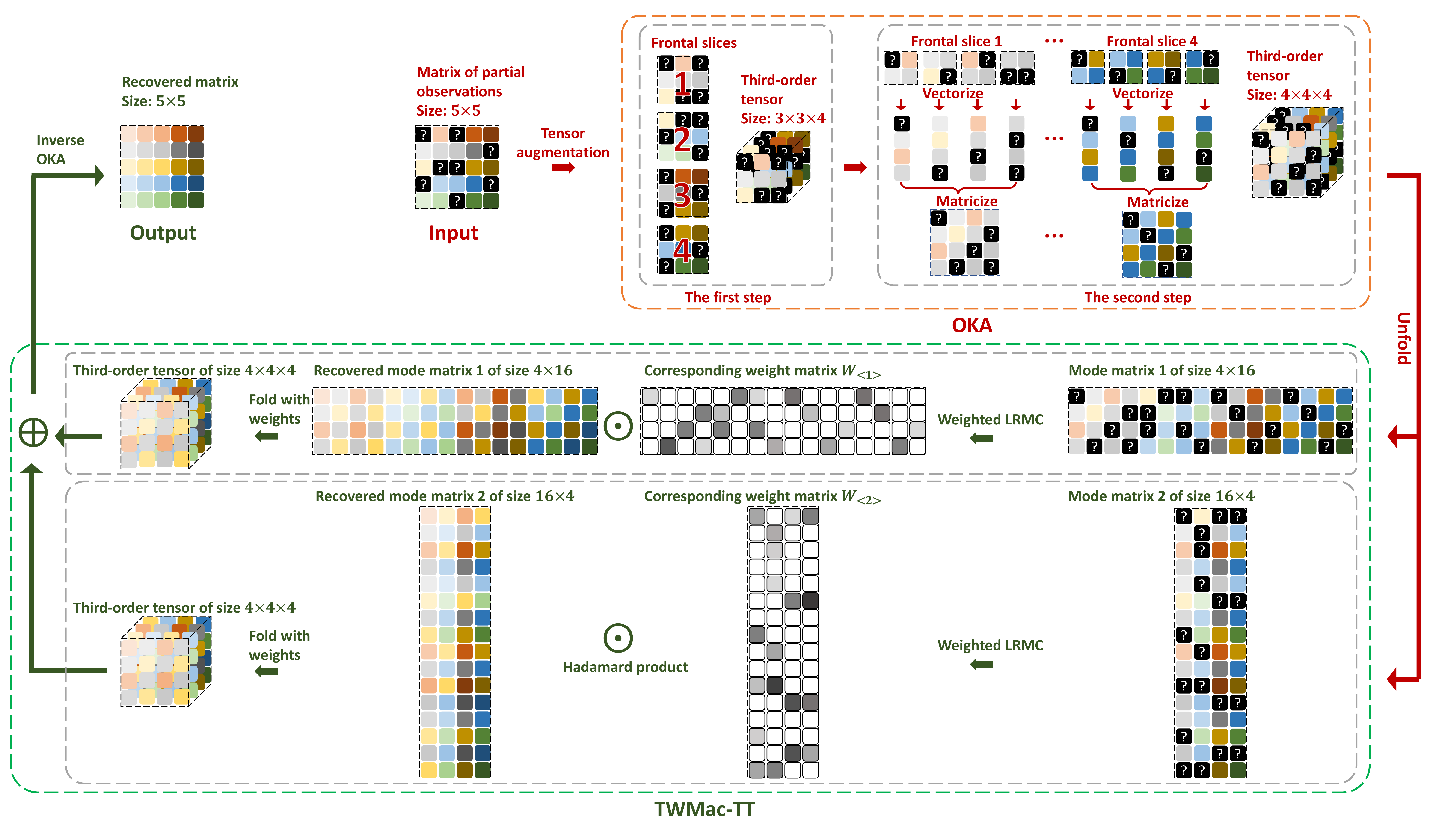} %
\caption{The pipeline of the proposed algorithm TWMac-TT-OKA in the form of a toy example, where the input is a $5\times 5$ matrix. The proposed algorithm mainly includes two stages, namely OKA and {TWMac-TT}. {In the pre-processing stage of tensor augmentation (the part boxed with a dotted orange rounded rectangle)}, the OKA uses two steps to {get a higher-order tensor} of size $4\times 4\times 4$. In the process of {TWMac-TT (the part boxed with a dotted green rounded rectangle)}, {the algorithm iteratively repeats the weighted low-rank} matrix decomposition until the algorithm converges.}
\label{flow}
\end{figure*}

In addition to the merit of the blocking artifact elimination, another advantage of OKA is that {OKA can deal with a tensor of arbitrary size.} {The tensor's size $[I_1, I_2, I_3]$ is no longer restricted by $I_1=I_2=2^n( n\in\mathbb N^{+})$, which is a limitation of KA.} Furthermore, we can deal with non-square tensors whose numbers of rows and columns are not equal. {Even though the dimension size of the input tensor is a prime number, we can still increase the order through the proposed overlapping mechanism.} 
In general, we can transform an arbitrary tensor to a higher-order one. We designed an automatic algorithm for computing the number of overlapped elements in every tensor augmentation step.

{We formulate this recursive algorithm into the general case, i.e.,}
\begin{equation}
\mathcal{T}_{\left[I_1\times I_2 \times I_3\right]}=\sum_{i_{p}, \ldots, i_{1}=1}^{4} \sum_{j=1}^{l} c_{i_{p} \cdots i_{1} j} \mathbf{e}_{i_{p}} \otimes \cdots \otimes \mathbf{e}_{i_{1}} \otimes \mathbf{u}_{j},
\end{equation}
where $p$ is determined by the input size and the number of overlapped elements. In Fig. \ref{OKAKA}(a), the number of overlapped elements is set to be 1 or 2, where the overlapping number {of each step} is determined by the size of the current tensor to be processed.  {We use $r, c$ to denote the number of rows and columns, respectively. If $r$ (or $c$) of the frontal slice is odd, the current overlapping elements in rows (or columns) will be 1. And if it is even, the overlapping number will be 2. That is, 
\begin{equation} \label{eq_over_illus}
    n_{\text{overlapping}} = 
    \begin{cases}
    1,& \text{if } r(\text{or } c)\bmod{2} = 1  \\
    2,& \text{if } r(\text{or } c)\bmod{2} = 0,
    \end{cases}
\end{equation}
where $\bmod$ is the modulo operation which represents the remainder of a division.
In practice, we find that the number of overlapping elements being 2 or 3 performs much better than others. Thus we overlap 2 or 3 elements in our experiments.
To be specific, if the number of rows (or columns) of the frontal slice is odd, the current overlapping number will be 3. And if it is even, the overlapping number will be 2, i.e.,
\begin{equation} \label{eq_over}
    n_{\text{overlapping}} = 
    \begin{cases}
    3,& \text{if } r(\text{or } c)\bmod{2}  = 1\\
    2,& \text{if } r(\text{or } c)\bmod{2}  = 0.
    \end{cases}
\end{equation}}

\begin{algorithm}
\caption{{The OKA procedure}}\label{alg:oka}{
\begin{algorithmic}
\STATE {\textbf{Input:}}The observed data $\mathcal{I}\in \mathbb{R}^{ \mathnormal{m \times n\times l}}$
\STATE{\textbf{Initialization:} The objective $Nway$ vector $[I_1, I_2, \ldots, I_N]$, the starting position index set $\{(x_{\text{start}}, y_{\text{start}})\}_{k=1}^{N-1}$, the frontal slice shape of each augmentation $\{(x_{\text{size}}, y_{\text{size}})\}_{k=1}^{N-1}$}
\FOR{ N-1 loops}
\STATE{$\mathcal{I}_{\text{old}} = \mathcal{I}$}
    \FOR{$k=1$ to Nway$(k)$}
        \STATE{$[i, j] = [x_{\text{start}}(k), y_{\text{start}}(k)]$}
        \STATE{$\mathcal{I}(:,i_k,:) = \mathcal{I}_{\text{old}}(:,i:i+x_{\text{size}}(k)-1, j:j+y_{\text{size}}(k)-1,:)$}
    \ENDFOR
\ENDFOR
\STATE{$\mathcal{T} = \mathcal{I}$}
\STATE {\textbf{Output:}}{ Higher-order tensor $\mathcal{T} \in \mathbb{R}^{I_1, I_2, \ldots, I_{N}}$}
\end{algorithmic}}
\end{algorithm}

\begin{algorithm}
    \caption{{The calculation for OKA initialization constants}}
    \label{alg_nway}{
    \begin{algorithmic}
        \STATE {\textbf{Iutput:}}{The observed data $\mathcal{I}\in \mathbb{R}^{ \mathnormal{m \times n\times l}}$ }
        \STATE{\textbf{Initialization:}  The input row length $r=m$, the input column length $c=n$, array $x_{\text{size}}(0) = r$, $y_{\text{size}}(0) = c$, $x_{\text{start}}=[]$, $y_{\text{start}}=[]$, the index $num=0$}
        \WHILE{r>4 \AND c>4}
            \STATE{$num = num+1$}
            \IF{$r\bmod{2}=1$ }
                \STATE{$r = (r+1)/2 +1 $}
            \ELSIF{$r\bmod{2}=0$}
                \STATE{$r = r/2 + 1$}
            \ENDIF
            \STATE{$x_{\text{size}}(num) = r$}
            \STATE{$x_{\text{start}}(num) = x_{\text{size}}(num-1) - x_{\text{size}}(num) +1$}
            \IF{$c\bmod{2}=1$}
                \STATE{$c = (c+1)/2 +1 $}
            \ELSIF{$c\bmod{2}=0$}
                \STATE{$c = c/2 + 1 $}
            \ENDIF
            \STATE{$y_{\text{size}}(num) = c$}
            \STATE{$y_{\text{start}}(num) = y_{\text{size}}(num-1) - y_{\text{size}}(num) +1$}
        \ENDWHILE
        \STATE{ Nway = concat($ones(1,num)\times 4, l)$}
        \STATE{$N = num+1$, $I_1 = I_2 = \ldots = I_{N-1} = 4, I_N = l$}
        \STATE {\textbf{Output:}} {The objective Nway vector $[I_1,I_2,\ldots,I_N]$, the starting position index set $\{(x_{\text{start}}, y_{\text{start}})\}_{k=1}^{N-1}$, and the frontal slice shape of each augmentation $\{(x_{\text{size}}, y_{\text{size}})\}_{k=1}^{N-1}$}
    \end{algorithmic}}
\end{algorithm}

Then, the {OKA illustrated in Fig. \ref{OKAKA} will keep repeating the structured block addressing procedure iteratively until the size of the resulting frontal slice of the tensor is equal to $4\times 4$. In other words, the tensor order reflected by the parameter $Nway$ keeps being increased until the tensor cannot be further divided. After obtaining the desired initialized parameter $Nway$ along with the start position $\{(x_{\text{start}}, y_{\text{start}})\}_{k=1}^{N-1})$ and the corresponding size of the divided block $\{(x_{\text{size}}, y_{\text{size}})\}_{k=1}^{N-1}$ for each division step,} the OKA procedure can be performed as described in Algorithm \ref{alg:oka}.

{The initialization constants in Algorithm \ref{alg:oka} are calculated by Algorithm \ref{alg_nway}. In Algorithm \ref{alg_nway}, the number of overlapped elements is calculated automatically via Eq. \eqref{eq_over}. Thus we can get the starting positions $\{(x_{\text{start}}, y_{\text{start}})\}_{k=1}^{N-1}$ and sizes $\{(x_{\text{size}}, y_{\text{size}})\}_{k=1}^{N-1}$ of sub-blocks at every division step $k$. After getting the objective $Nway$ and division scheme indicated by the starting positions and sizes of each sub-block, we augment the observed low-order tensor $\mathcal{I}$ into the higher-order one ${\mathcal{T}}$ using Algorithm \ref{alg:oka}.}

\subsection{TWMac-TT-OKA Algorithm}
To solve the weighted model EWLRTC-TT in \eqref{objectfunctionours}, we take its partial derivative with respect to {$U_k$ and $V_k$}, which gives 
{
\begin{equation}
\frac{\partial J}{\partial U_k} = 2(W_k\odot(U_k {V_k}'-\mathcal{X}_{<k>}))V_k,
\end{equation}
\begin{equation}
\frac{\partial J}{\partial V_k} = 2(W_k\odot(V_k {U_k}'-{\mathcal{X}_{<k>}}'))U_k.
\end{equation}
}

To ensure that the solution is identifiable, the $\ell_2$ norm penalty for both {$U_k$ and $ V_k$} is introduced into the above formulation:
{
\begin{equation}
\min_{U_k, V_k} ||W_k\odot (\mathcal{X}_{<k>}-U_k {V_k}') ||_p+\frac{\lambda_u}{2}|| U_k||_2^2+\frac{\lambda_v}{2}||V_k||_2^2.
\label{solution}
\end{equation}
}Thus, it is easy to get a closed-form formula by setting the partial derivative to zero, i.e., 
{
\begin{equation}\label{V}
V_{kj} = ({\hat U_k}'\Omega_j {\hat U_k} + \lambda_v I_r)^{-1}{\hat U_k}'\Omega_j \mathcal{X}_{<k>_j},
\end{equation}
\begin{equation} \label{U}
U_{ki} = ({\hat V_k}'\Lambda_i {\hat V_k} + \lambda_u I_r)^{-1}{\hat V_k}'\Lambda_i \mathcal{X}_{<k>i},
\end{equation}}where  $\Omega_j \in \mathbb R^{m\times m}$ is a diagonal matrix with the elements from the $j$-th column of {$\mathcal{X}_{<k>}$} and $\Lambda_i \in \mathbb R^{n\times n}$ is a diagonal matrix with the elements from the $i$-th row of {$\mathcal{X}_{<k>}$}.

Since  the $\ell_2$ norm penalty is used (\ref{solution}), we update the weight {$W_k$ for the $k$-th mode} via a convex function following the equation used in {\cite{guo2015generalized}}: 
{
\begin{equation} \label{W}
W_k = c\sqrt{exp(-\gamma|\mathcal{X}_{<k>}-U_k{V_k}'|)},
\end{equation}}where {hyper-parameters} $c$ and $\gamma$ are positive constants. As a result, by iteratively calculating {$U_k, V_k$ and $W_k$}, we can guarantee a (local) optimal solution {\cite{icml03}}.

We apply the block coodinate descent (BCD) algorithm, following TMac and TC-MLFM used in \cite{xu2013parallel} and \cite{tan2014tensor}. More precisely,  after updating {$U_{k}^{l+1}$, $V_{k}^{l+1}$ and $W_{k}^{l+1}$ for all $k=1,2,\ldots,N-1$}, we compute the elements of the tensor $\mathcal{X}^{l+1}$ as follows:
{
\begin{equation}\label{fold}
x_{i_1 \cdots} = \left\{
\begin{array}{ll}
\Bigl(\sum_{k=1}^{N-1}\text{fold}_{\mathcal{W}}(W_{k}\odot \mathcal{X}_{<k>})\Bigr)_{i_1\cdots} & (i_1\cdots) \notin \Omega\\
t_{i_1\cdots} & (i_1\cdots) \in \Omega \\
\end{array} \right .,
\end{equation}}where $\text{fold}_\mathcal W$ is a fold operation which folds all the mode matrices according to their element-wise weights. This algorithm is described as tensor completion by parallel weighted matrix factorization based on tensor train with overlapping ket augmentation (TWMac-TT-OKA). The detailed algorithm description is summarized in Algorithm \ref{alg:alg3}. 

\begin{algorithm}[H]{
\caption{{TWMac-TT-OKA}}\label{alg:alg3}
\begin{algorithmic}
\STATE {\textbf{Input:}}The observed data $\mathcal{I}\in \mathbb{R}^{ \mathnormal{m \times n \times l}}$, index set $\Omega$
\STATE {\textbf{Pre-processing:}} Augment the input tensor by OKA algorithm and get $\mathcal{T}\in \mathbb{R}^{ \mathnormal{I_1 \times I_2 \cdots \times I_N}}$
\STATE {\textbf{Parameters:}} $th, c, \gamma, W_{i}$ for $i = 1,\dots,N-1$
\STATE {\textbf{Initialization:}} $U_k^0,V_k^0,\mathcal{X}^0$ with $\mathcal{X}_\Omega^0 = \mathcal{T}_\Omega, l=0$
\STATE {\textbf{while not converged do:}}
\STATE \hspace{0.3cm}\textbf{for} $k=1$ to $N-1$ \textbf{do}
\STATE \hspace{0.6cm} Unfold the tensor $\mathcal{X}^l$ to get $\mathcal{X}_{<k>}^l$
\STATE \hspace{0.6cm} \textbf{for} $j=1$ to \# columns of $\mathcal{X}_{<k>}^l$ \textbf{do}
\STATE \hspace{0.9cm} $V_{kj}^{l+1}=((\hat U_k^{l})'\Omega_j^{l} \hat U_k^{l}+ \lambda_v I_r)^{-1} (\hat U_k^{l})'\Omega_j^{l} \mathcal{X}_{<k>j}^{l}$
\STATE \hspace{0.6cm} \textbf{end for}
\STATE \hspace{0.6cm} \textbf{for} $i=1$ to \# rows of $\mathcal{X}_{<k>}^l$ \textbf{do}
\STATE \hspace{0.9cm} $U_{ki}^{l+1} = ((\hat V_k^{l+1})'\Lambda_i^l \hat V_k^{l+1} + \lambda_u I_r)^{-1}(\hat V_k^{l+1})'\Lambda_i^l \mathcal{X}_{<k>i}^{l}$
\STATE \hspace{0.6cm} \textbf{end for}
\STATE \hspace{0.6cm} $W_{k}^{l+1} = c\sqrt{exp(-\gamma|\mathcal{X}_{<k>}^{l}-U_k^{l+1} (V_k^{l+1})'|)}$
\STATE \hspace{0.3cm} \textbf{end for}
\STATE Update the tensor $\mathcal{X}^{l+1}$ using \\
$x_{i_1 \cdots}^{l+1} = \left\{
\begin{array}{ll}
\Bigl(\sum_{k=1}^{N-1}\text{fold}_{\mathcal W}(W_{k}^{l+1}\odot \mathcal{X}_{<k>}^{l+1})\Bigr)_{i_1\cdots} & (i_1\cdots) \notin \Omega\\
t_{i_1\cdots} & (i_1\cdots) \in \Omega \\
\end{array} \right .$\\%
\STATE \textbf{end while}
\STATE {\textbf{Output:}} The recovered tensor of its original order $\hat{\mathcal{I}}\in\mathbb{R}^{m\times n\times l}$
\end{algorithmic}}
\end{algorithm}

To make the process clearer, we use an example to display the pipeline of the proposed algorithm in Fig. \ref{flow}. We assume that there is a $5\times 5$ matrix with partial observations. The missing elements are represented by the black square containing a question mark. To recover the original matrix, TWMac-TT-OKA mainly carries out two stages, boxed with {dotted} rounded rectangles in orange and green, respectively. { The first stage is the pre-processing, which }aims to augment the input via the OKA scheme. In particular, the example matrix does not meet the conditions of KA and reshaping, as 5 is a prime number and can be divided by only one and itself. {In this specific case, among OKA, KA and reshape, only OKA can increase the order of the matrix, as OKA can deal with matrices or tensors of arbitrary size.} {This pre-processing procedure in the rounded orange rectangle is similar to the procedure in Fig. \ref{OKAKA}}. In the second phase, TWMac-TT is applied to the augmented tensor of size $4\times 4 \times 4$. Two mode matrices are formed by unfolding the tensor using mode-$i$ canonical matricization, i.e., the mode matrix 1 of size $4\times 16$ and the mode matrix 2 of size $16\times 4$. These two matrices are then completed using the weighted LRMC to acquire the weight {matrix estimation and the factor matrix estimations}. {The weight matrix is a gray-scale matrix, where the values at the known locations are one, and the values at unknown locations are filled with estimated weights ranging from 0 to 1.} {Then, we can obtain two third-order tensors by multiplying the weight matrices and the mode matrices with the Hadamard product and folding them with the estimated element-wise weights.} We add these two tensors together to get a recovered tensor. The process in the second phase is repeated until the algorithm converges. By conducting the inverse operation of OKA, {we recover the given incomplete tensor ${\mathcal{I}}$ as the output $\hat{\mathcal{I}}$ of the proposed algorithm.}

\section{Experiments} \label{exp}
We conduct extensive experiments on synthetic data, real color images and magnetic resonance imaging (MRI) data to demonstrate the effectiveness of our model and algorithm. 
We compare our {TWMac-TT-OKA} algorithm with several classic and state-of-the-art tensor completion methods, including TMac \cite{xu2013parallel}, SiLRTC \cite{liu2013tensor}, FBCP \cite{zhao2015bayesian}, STDC \cite{chen2013simultaneous}, and TMac-TT \cite{bengua2017efficient}.

The proposed methods are {TMac-TT+OKA, TWMac-TT and TWMac-TT+OKA}, among which TWMac-TT+OKA is our {final} model.


\subsection{Synthetic data completion}
\label{sec_syn}

We conduct a series of simulations to achieve two {primary} goals. The first goal is to validate the effectiveness of weight estimation of the proposed model and {algorithm}. The second goal is to demonstrate the superiority of our method over other compared methods. The simulated low-TT-rank tensor $\mathcal{X}\in \mathbb{R}^{I_1\times I_2\cdots \times I_N}$ is generated simply by using the TT representation formula\cite{oseledets2011tensor}:
\begin{equation}
\mathcal{I}(i_1,i_2,\dots,i_N) = \sum_{\alpha = 1}^{r} U_1(i_1,\alpha)U_2(i_2,\alpha)\dots U_N(i_N,\alpha),
\end{equation}
where the decomposition components $U_i \in \mathbb{R}^{I_i\times r_i}$ are generated randomly according to a standard Gaussian distribution, i.e., {$U_i \sim \mathcal{N}(0,1)$}. For convenience, the dimension of each TT mode and the corresponding TT ranks are set equally as $I_1 = I_2 = \cdots = I_N = I$ and $r_1 = r_2 = \cdots = r_{N-1} = r$, respectively.
\begin{figure}[htb]
\centering
\subfigure[Iteration 2]{\includegraphics[width=0.2\textwidth]{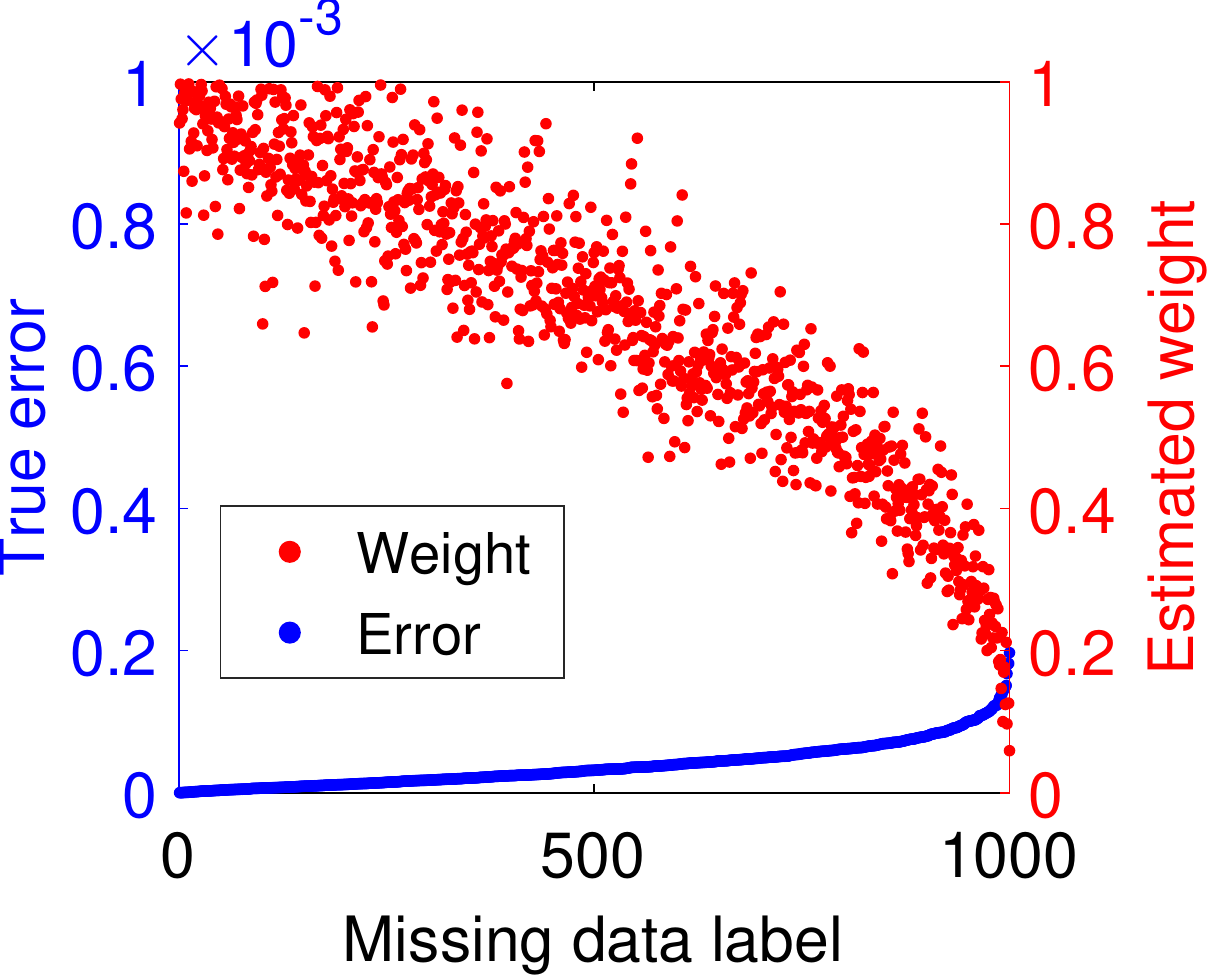}}
\subfigure[Iteration 4]{\includegraphics[width=0.2\textwidth]{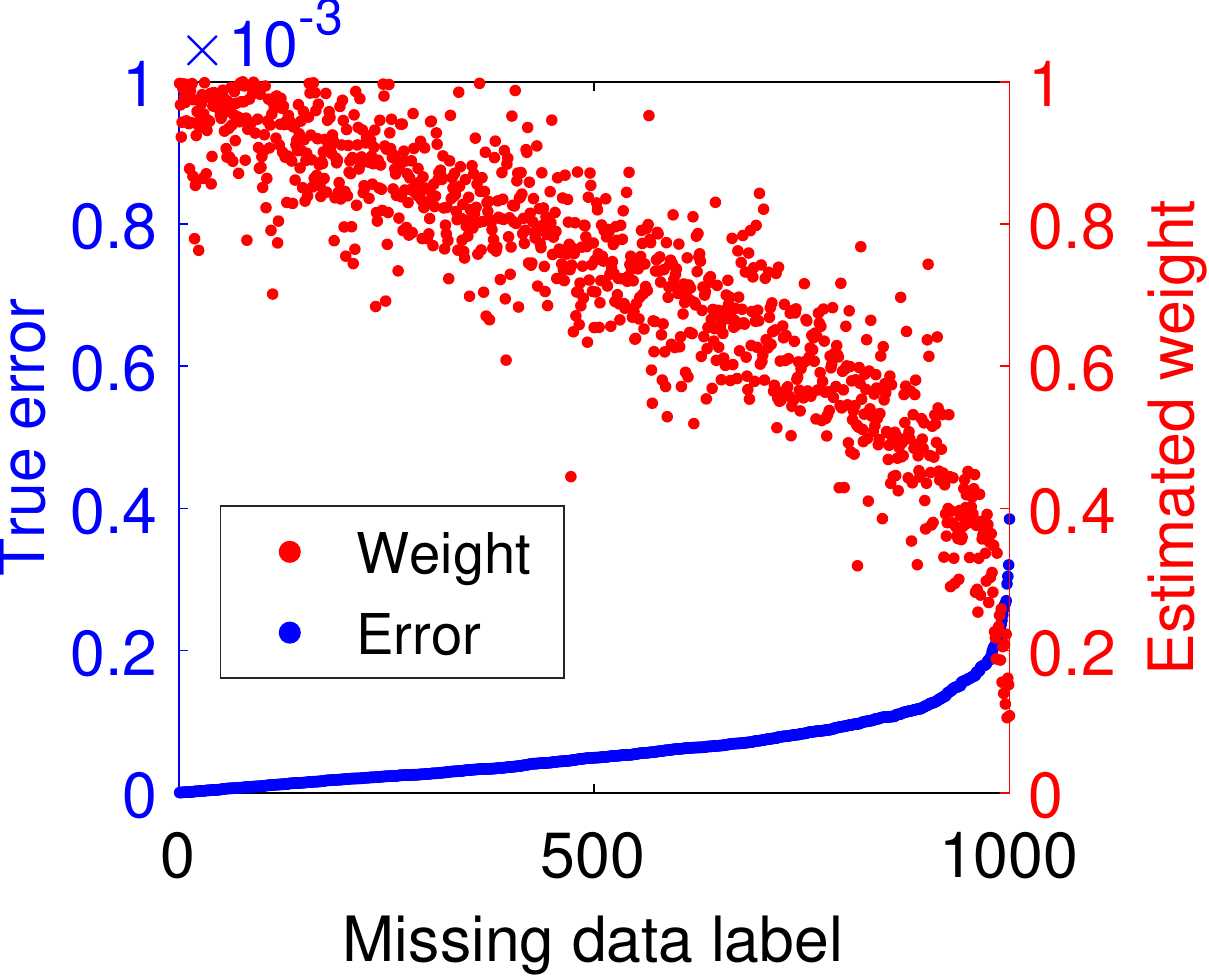}}
\subfigure[Iteration 6]{\includegraphics[width=0.2\textwidth]{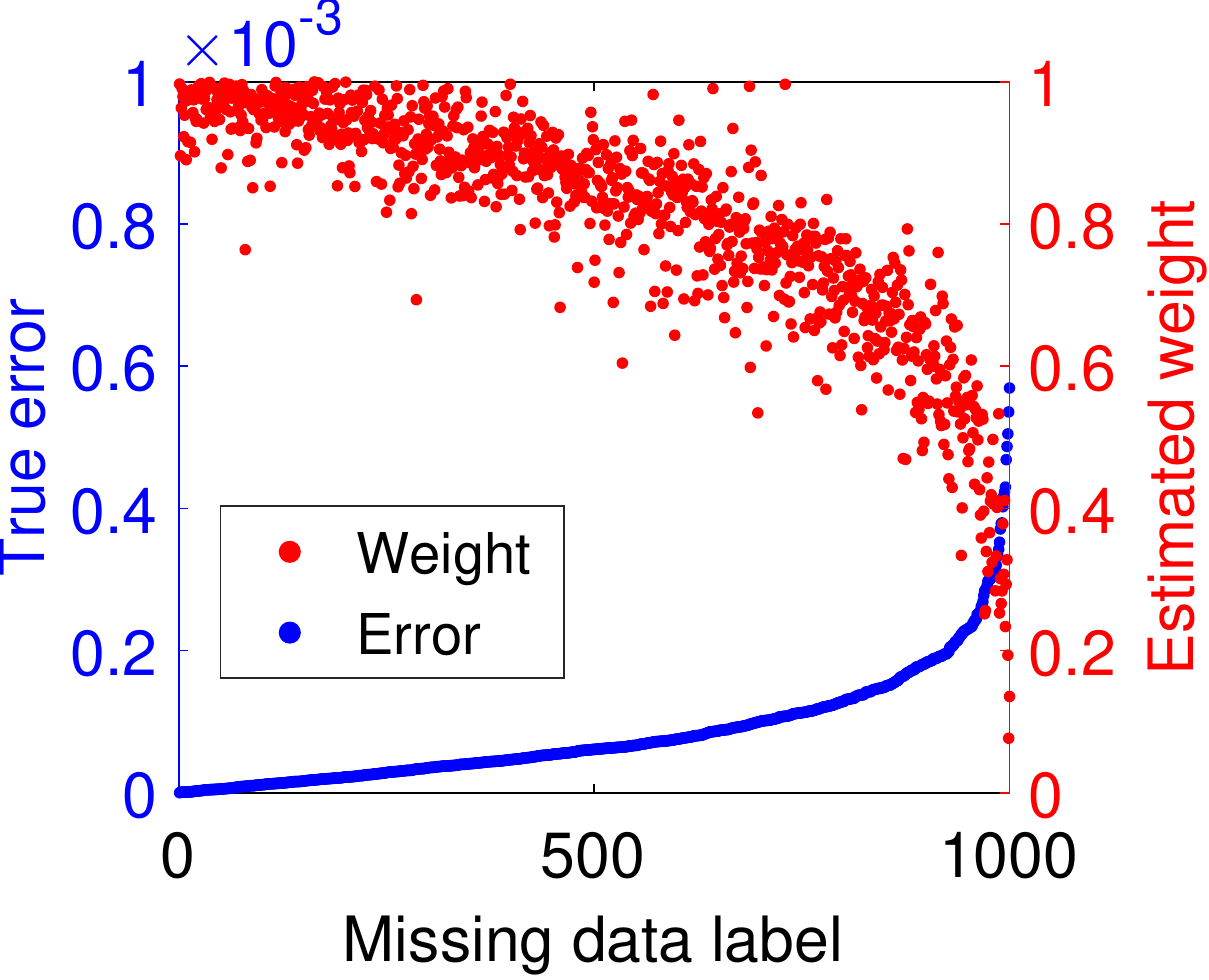}}
\subfigure[Iteration 8]{\includegraphics[width=0.2\textwidth]{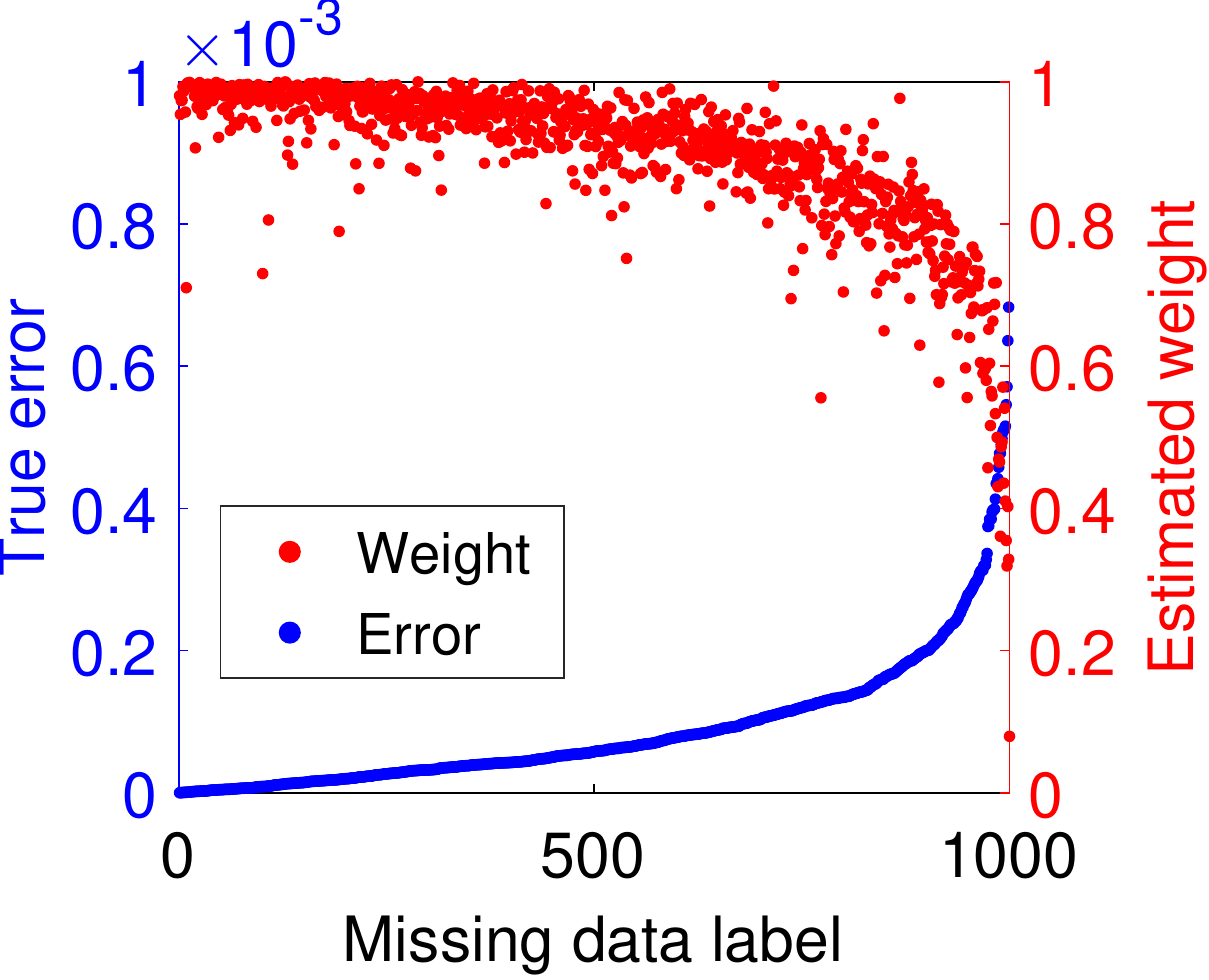}}
\caption{The scatter diagram of true errors and corresponding estimated weights with missing rate of 50 percent.}
\label{error and weith with 50 percent}
\end{figure}

\begin{figure}[htb]
\centering
\subfigure[Iteration 8]{\includegraphics[width=0.2\textwidth]{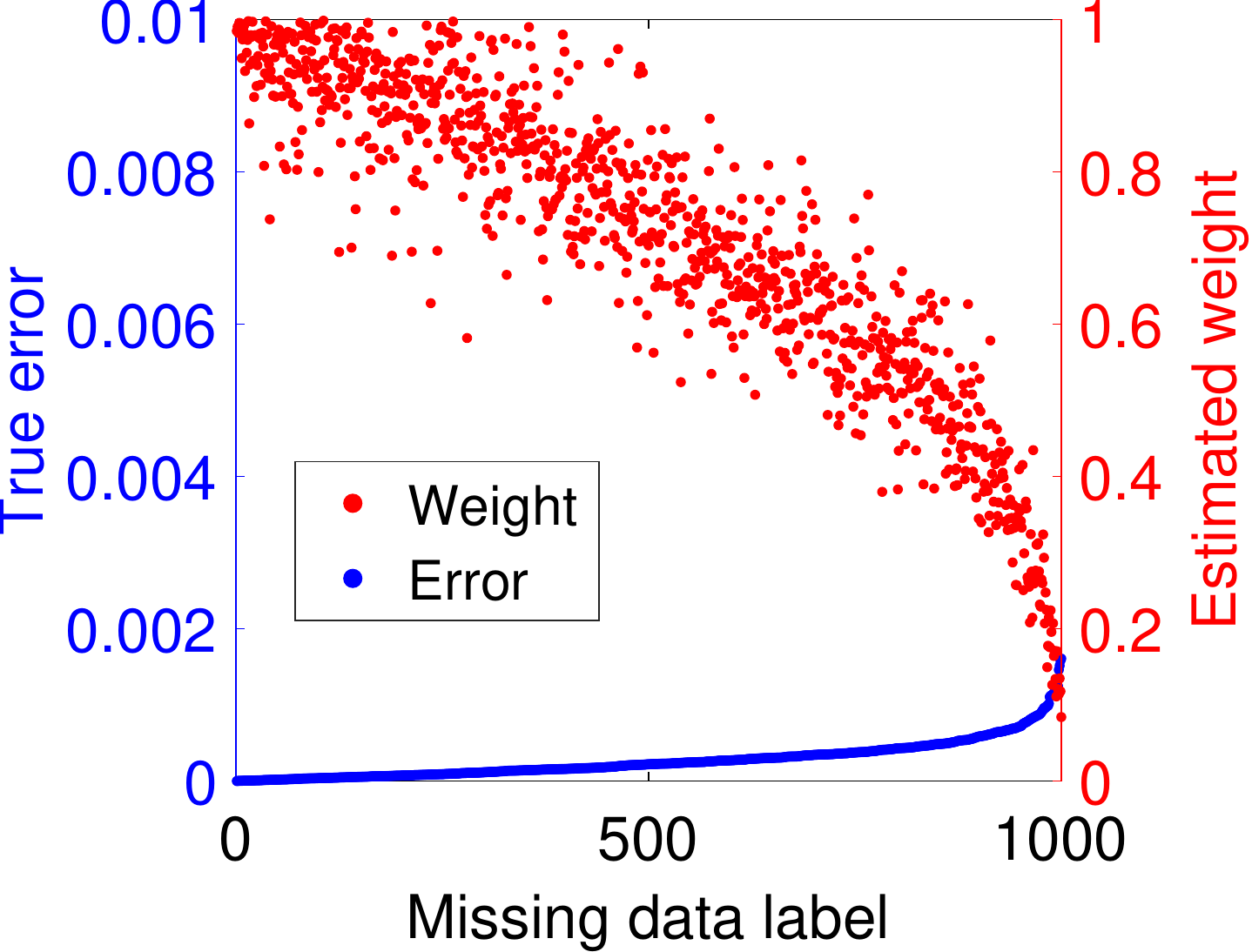}}
\subfigure[Iteration 10]{\includegraphics[width=0.2\textwidth]{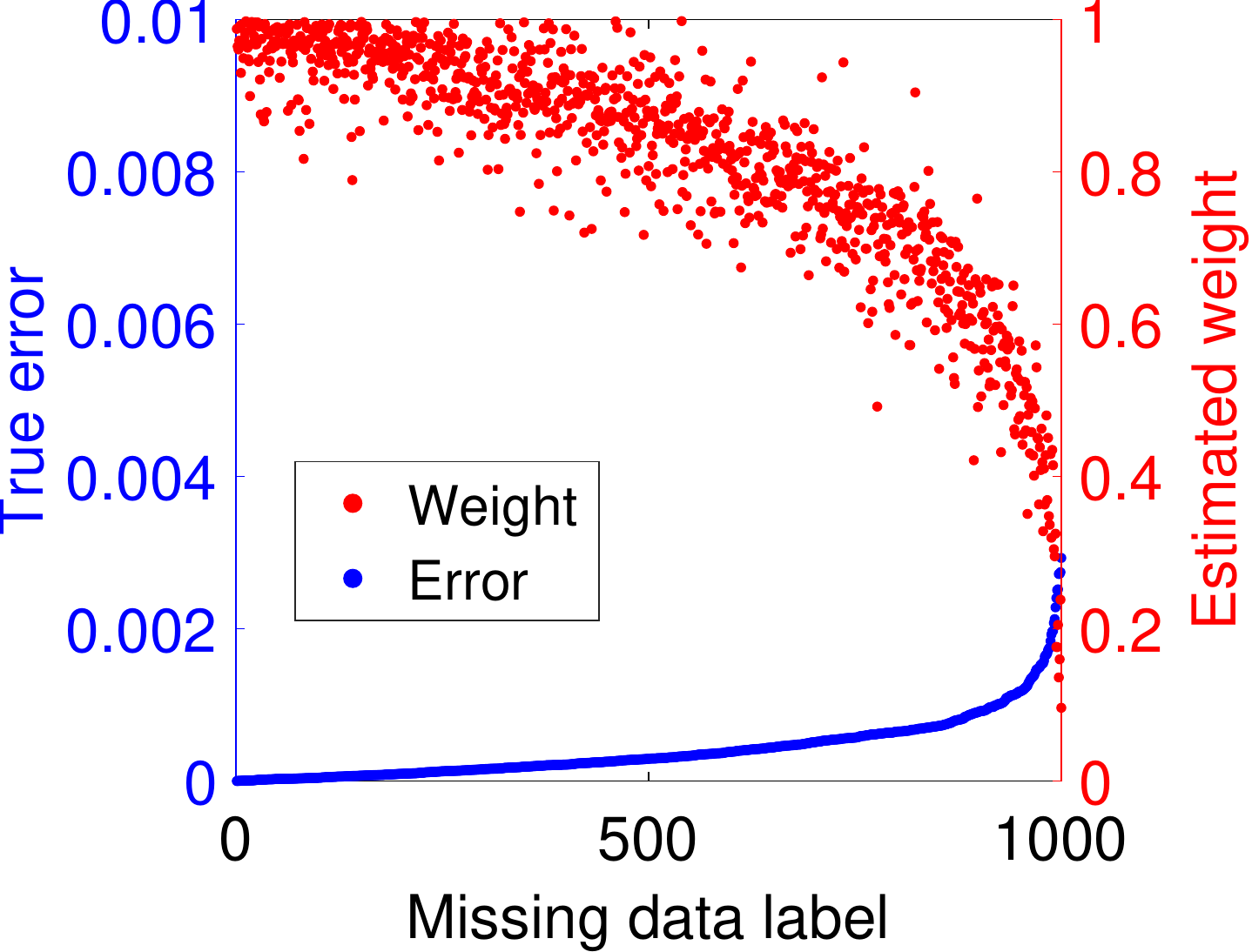}}
\subfigure[Iteration 12]{\includegraphics[width=0.2\textwidth]{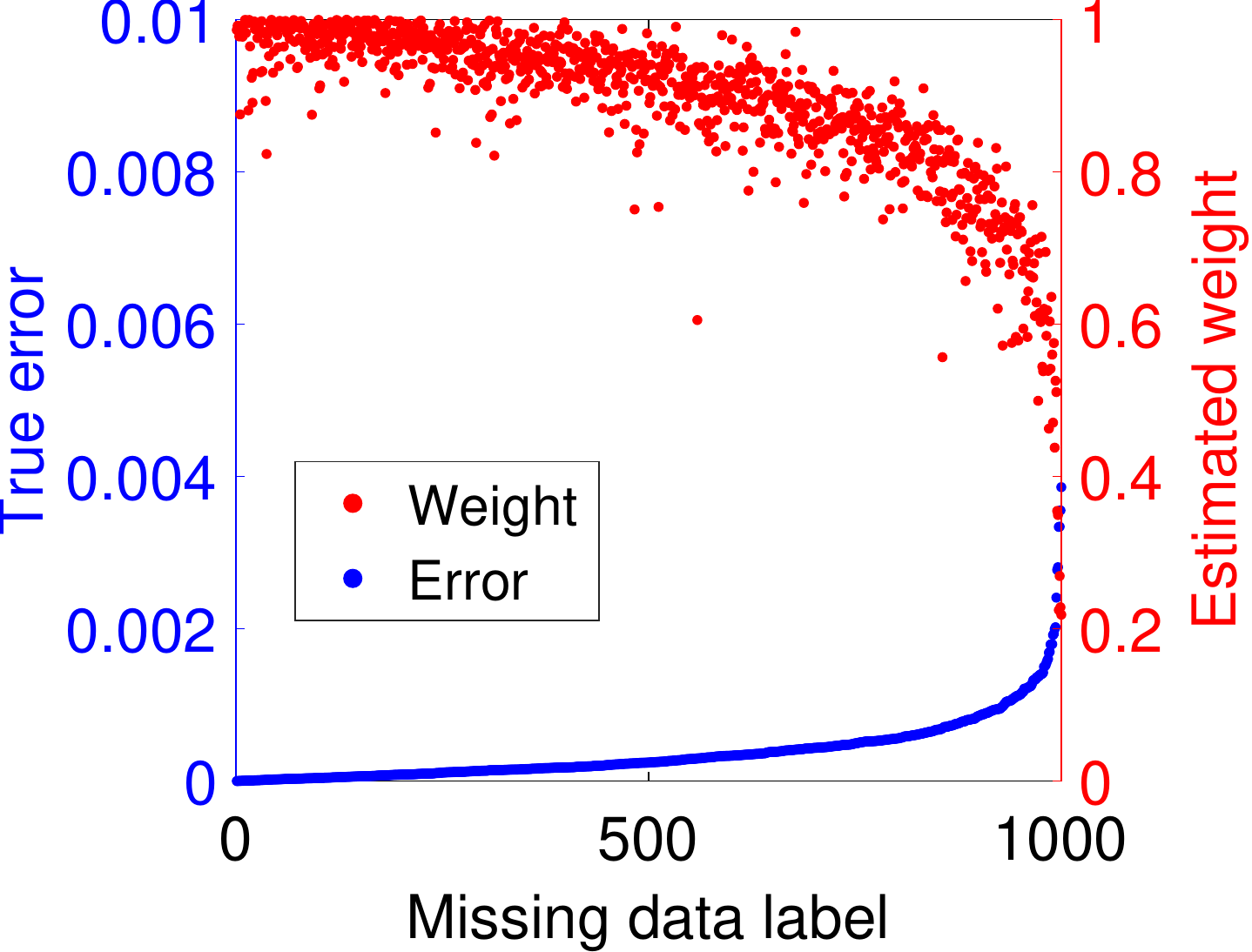}}
\subfigure[Iteration 14]{\includegraphics[width=0.2\textwidth]{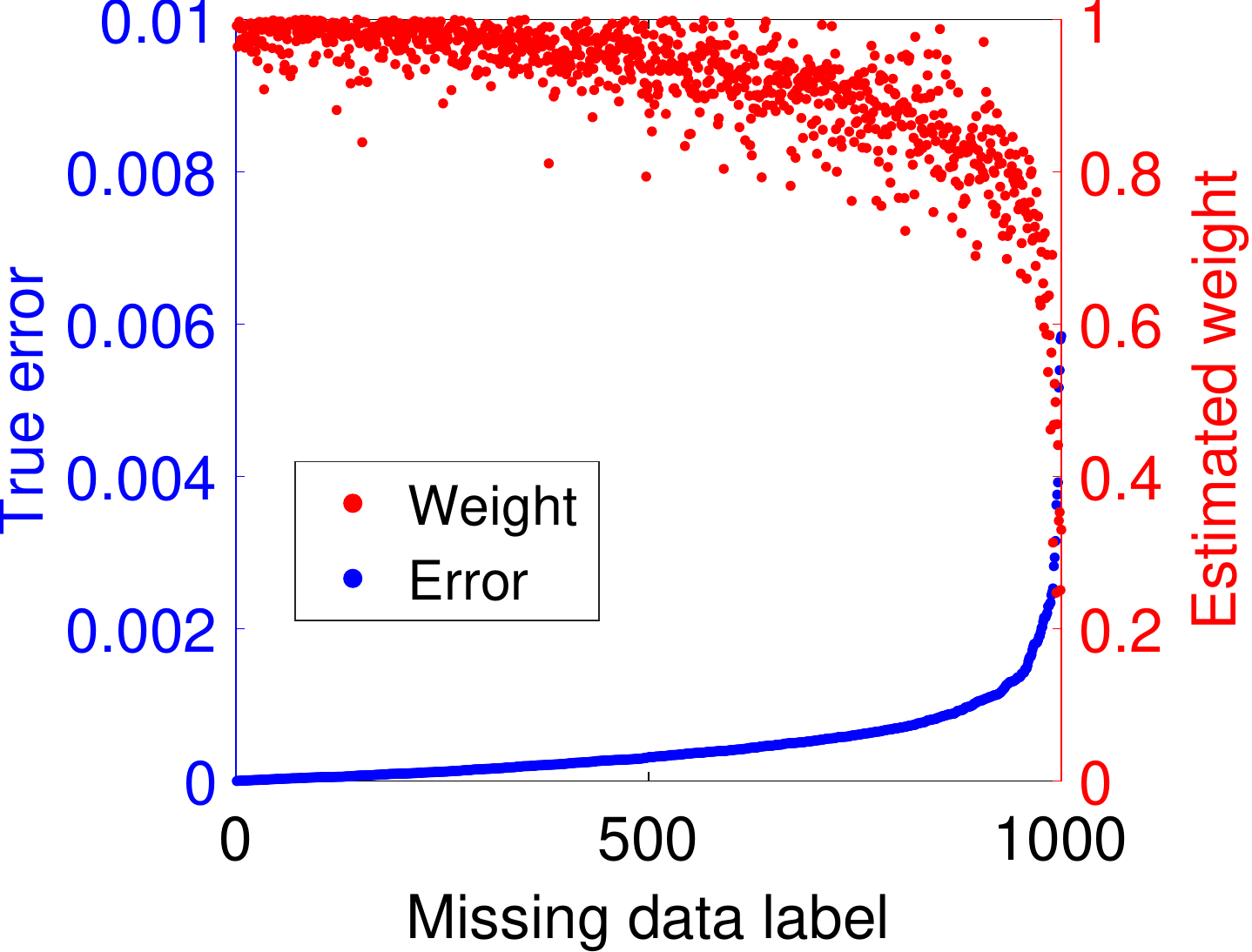}}
\caption{The scatter diagram of true errors and corresponding estimated weights with missing rate of 70 percent.}
\label{error and weith with 70 percent}
\end{figure}

In general, we conduct four sets of experiments for {tensors of} different sizes to cover the different tensor orders and ranks, including a fourth-order tensor {$\mathcal{I}\in \mathbb{R}^{40\times40\times40\times40}$}, a fifth-order tensor {$\mathcal{I}\in \mathbb{R}^{20\times20\times20\times20\times20}$}, a sixth-order tensor {$\mathcal{I}\in \mathbb{R}^{10\times10\times10\times10\times10\times10}$} and a seventh-order tensor {$\mathcal{I}\in \mathbb{R}^{10\times10\times10\times10\times10\times10\times10}$}. The corresponding TT ranks are set as (10,10,10) (fourth-order), (5,5,5,5) (fifth-order), (4,4,4,4,4) (sixth-order) , and (4,4,4,4,4,4) (seventh-order), respectively.

To validate the quality of the weight estimation procedure for the proposed TWMac-TT algorithm, we plot the scatter diagrams of the true errors and the estimated weights obtained by the proposed algorithm for a fourth-order synthesized tensor with different missing rates, namely, 50\% and 70\%. More details are shown in Fig. \ref{error and weith with 50 percent}-\ref{error and weith with 70 percent}.

We randomly choose 1000 elements from the missing elements for illustration {in these two instances}. {We can intuitively assess the relationship between the estimated weights and the actual recovery errors. } As seen, both {scatter plots} under different missing rates show an inverse relation between the weights and recovery errors, i.e., larger recovery errors correspond to smaller estimated weights, and vice versa, which evidences that the weight estimation in the proposed algorithm is accurate. Furthermore, from the perspective of the iterations of this experiment, as the errors decrease, the estimated weights get correspondingly larger, and the profiles of the curves of the weights {(red ones in the plots)} are simultaneously getting "thinner", which means that the estimation errors are getting smaller. These findings verify the effectiveness of the element-wise weight estimation, which is the basis of the validity of our model, and demonstrate the convergence of the proposed algorithm. 

\begin{figure}[htb]
\centering
\includegraphics[width=0.4\textwidth]{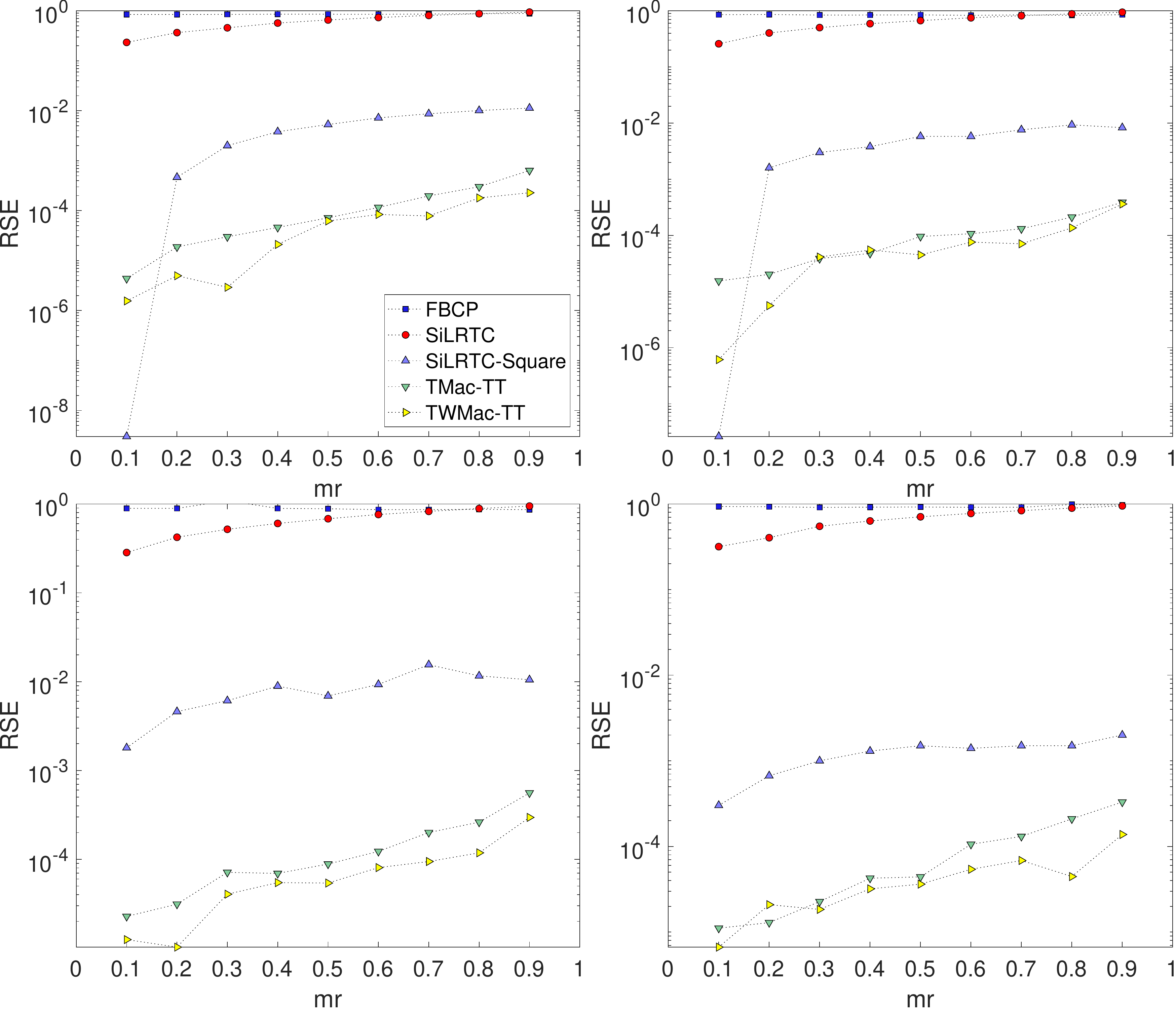} 
\caption{The RSE comparison between different LRTC algorithms for different sizes of synthetic tensors. From top to bottom, left to right, they are results of different dimensions 4D, 5D, 6D, and 7D, respectively.}
\label{linechart}
\end{figure}

We then compare our algorithm with others in terms of the RSE in Fig. \ref{linechart}. In Fig. \ref{linechart}, different settings of input tensors are evaluated. From top to bottom, left to right, they are the results of dimensions 4D, 5D, 6D, and 7D, respectively. We can see from these plots that TWMac-TT performs the best in most cases, especially where the missing rates are large, e.g., $mr = 0.9$. Among all the compared algorithms, FBCP has the worst performance. Compared with the baseline TMac-TT, TWMac-TT achieves a gain performance by a large margin.

\subsection{Color image completion}\label{color_sec}
Five color images are employed for the evaluation: Lena, Peppers, Sailboat, Baboon and Airplane. These images are represented as third-order tensors of size $256\times 256\times 3$. Due to the FBCP, SiLRTC, STDC, and TMac being based on the original tensor form, the input of these methods is the RGB third-order tensor. {Correspondingly, the tensor needs to be pre-processed into a higher-order one for TMac-TT+KA, TMac-TT+RE, and the proposed TWMac-TT+OKA. } KA increases the order of the tensors to nine that are sized $4\times4 \times4 \times4 \times4\times4 \times4\times4 \times3$, and so does the reshape operation. {Due to the overlap operation, the order of tensors can be further increased for the proposed OKA procedure.}  We set the number of overlapped pixels {to} be 2 and 3 {based on Eq. \eqref{eq_over}}. {The output size is automatically calculated by Algorithm \ref{alg_nway}}. Thus, the tensor transformed by OKA is a tenth-order tensor of size $4\times4\times4\times4\times4\times4\times4\times4\times4\times3$.

\begin{figure*}
  \centering
    \begin{minipage}[b]{1\textwidth}\centering{
      {\includegraphics[width=0.6in]{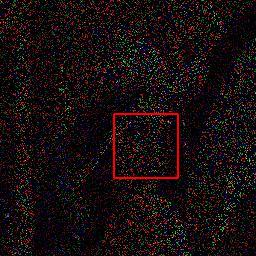}}
      {\includegraphics[width=0.6in]{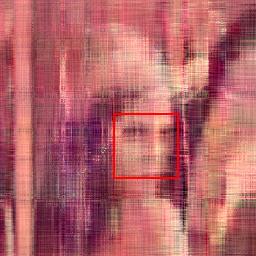}}
      {\includegraphics[width=0.6in]{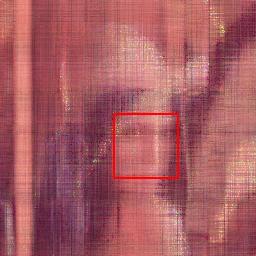}}
      {\includegraphics[width=0.6in]{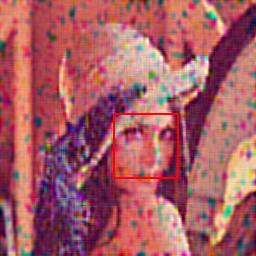}}
      {\includegraphics[width=0.6in]{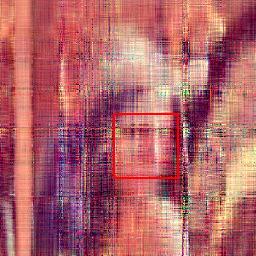}}
      {\includegraphics[width=0.6in]{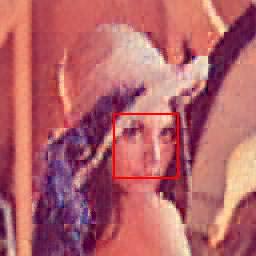}}
      {\includegraphics[width=0.6in]{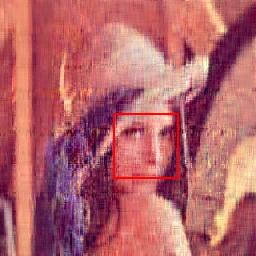}}
      {\includegraphics[width=0.6in]{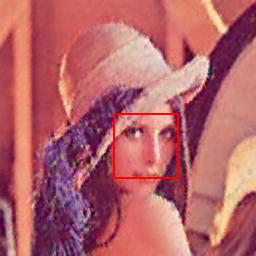}}
      {\includegraphics[width=0.6in]{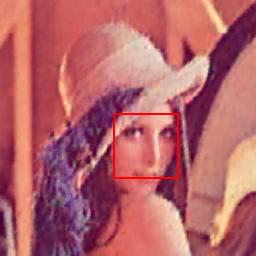}}
      }
    \end{minipage}\vspace{0.03in}
    \begin{minipage}[b]{1\textwidth}\centering{
      {\includegraphics[width=0.6in]{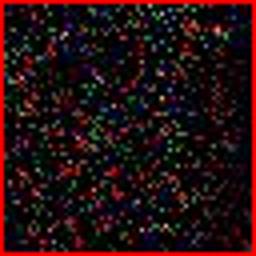}}
      {\includegraphics[width=0.6in]{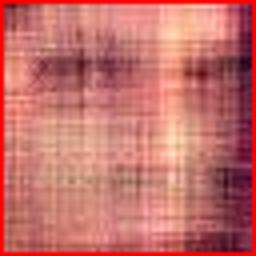}}
      {\includegraphics[width=0.6in]{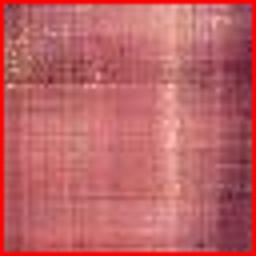}}
      {\includegraphics[width=0.6in]{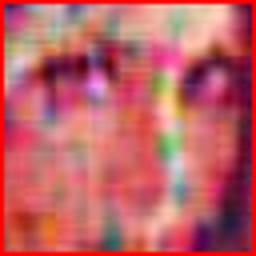}}
      {\includegraphics[width=0.6in]{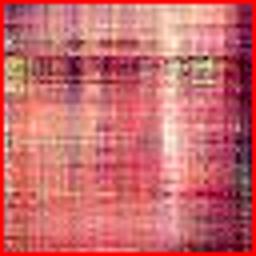}}
      {\includegraphics[width=0.6in]{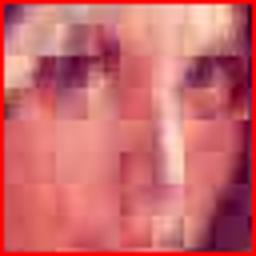}}
      {\includegraphics[width=0.6in]{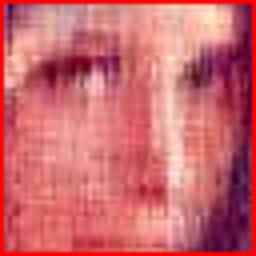}}
      {\includegraphics[width=0.6in]{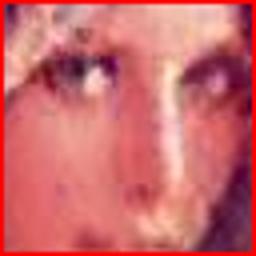}}
      {\includegraphics[width=0.6in]{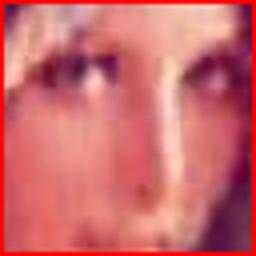}}
      }
    \end{minipage}\vspace{0.03in}

  \begin{minipage}[b]{1\textwidth}\centering{
      {\includegraphics[width=0.6in]{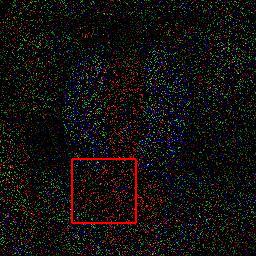}}
      {\includegraphics[width=0.6in]{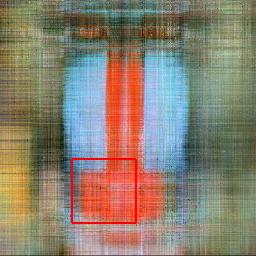}}
      {\includegraphics[width=0.6in]{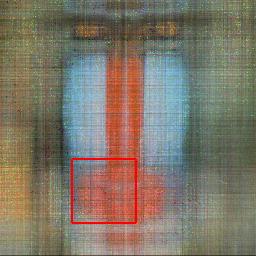}}
      {\includegraphics[width=0.6in]{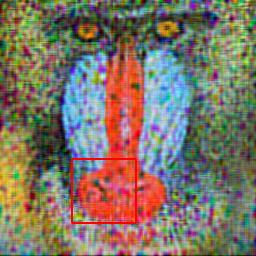}}
      {\includegraphics[width=0.6in]{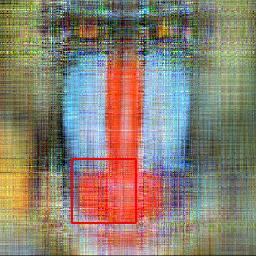}}
      {\includegraphics[width=0.6in]{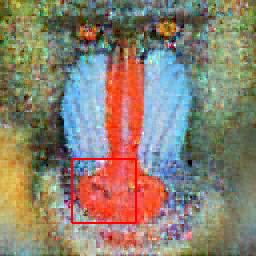}}
      {\includegraphics[width=0.6in]{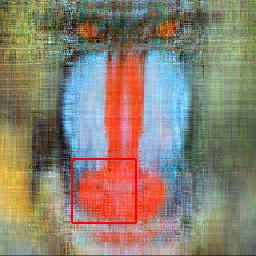}}
      {\includegraphics[width=0.6in]{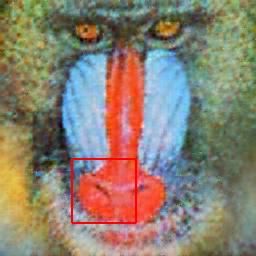}}
      {\includegraphics[width=0.6in]{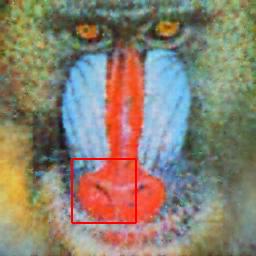}}
  }
  \end{minipage}\vspace{0.03in}
  \begin{minipage}[b]{1\textwidth}\centering{
      {\includegraphics[width=0.6in]{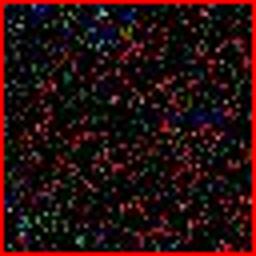}}
      {\includegraphics[width=0.6in]{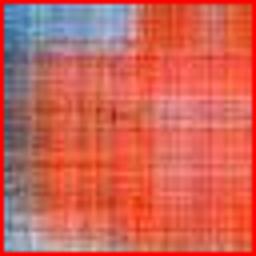}}
      {\includegraphics[width=0.6in]{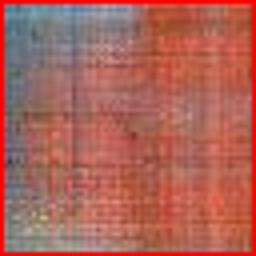}}
      {\includegraphics[width=0.6in]{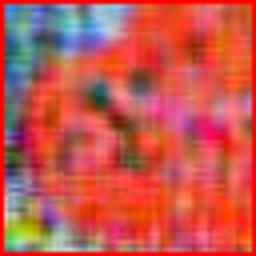}}
      {\includegraphics[width=0.6in]{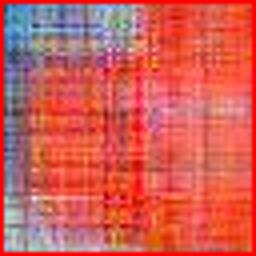}}
      {\includegraphics[width=0.6in]{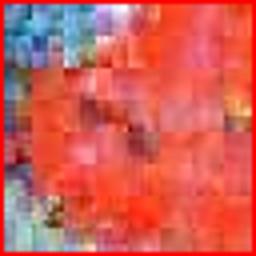}}
      {\includegraphics[width=0.6in]{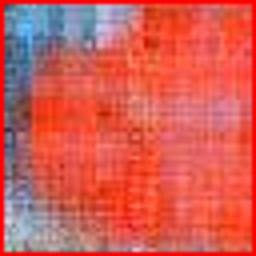}}
      {\includegraphics[width=0.6in]{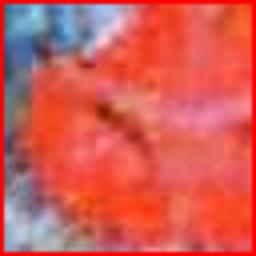}}
      {\includegraphics[width=0.6in]{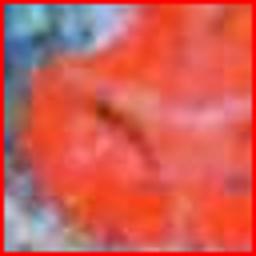}}
  }
  \end{minipage}\vspace{0.03in}

  \caption{Completion results of Lena and Baboon with 90\% missing elements. (a) Missing, (b) FBCP, (c) SiLRTC, (d) STDC, (e) TMac, (f) TMac-TT+KA, (g) TMac-TT+RE, (h) TMac-TT+OKA, and (i) TWMac-TT+OKA. {The figures in the last two columns are obtained by the proposed algorithm. For the convenience of comparison, each group of images is partially enlarged below the corresponding original-size results.}}
  \label{color}
\end{figure*}

\renewcommand{\arraystretch}{1}
\begin{table*}[htp]
 \caption{The average recovery performance (RSE, PSNR, SSIM) on five images
with missing ratios of 50, 60, 70, 80 and 90 percent.}
 \label{tab:rgb}
  \centering
 \fontsize{7}{8}\selectfont
  \begin{threeparttable}
  \label{tab:performance_comparison}
    \begin{tabular}{p{1cm}p{1cm}p{1cm}p{1cm}p{1cm}p{1cm}p{1cm}p{1cm}p{1cm}p{1cm}p{1cm}}
    \toprule
    & &FBCP&SiLRTC&STDC&TMac&TMac-TT+KA&TMac-TT+RE&TMac-TT+OKA&TWMac-TT+OKA
    \cr
    \midrule
    \multirow{3}{*}{50\%}
    & RSE    &0.0753 &0.0745 &0.0717 &0.1086  &0.0791 &0.0643  &\underline{0.0487} &\textbf{0.0434}\cr
    & PSNR   &27.3263&27.3671&27.7361&23.8824 &26.6803&29.0196 &\underline{31.6058}&\textbf{32.5480}\cr
    & SSIM   &0.8541 &0.8795 &0.8728 &0.7091  &0.8973 &0.8834  &\underline{0.9502} &\textbf{0.9594}\cr
  	\hline

	\multirow{3}{*}{60\%}
	& RSE   &	0.0908 &0.0927 & 0.0797&0.1164  &0.0905  &0.0817  &\underline{0.0579} &\textbf{0.0526}\cr
	& PSNR	& 25.6578  &25.4150&26.8793&23.2763 &25.6455 &26.7737 &\underline{30.0204}&\textbf{30.7904}\cr
	& SSIM  &	0.7934 &0.8173 &0.8457 &0.6714  &0.8453  &0.8238  &\underline{0.9277} &\textbf{0.9387}\cr
    \hline

    \multirow{3}{*}{70\%}
    & RSE   &0.1106 &	0.1179 &0.0934 &0.1273  &0.1001 &0.1004 &\underline{0.0731} &\textbf{0.0688}\cr
	& PSNR	&23.8795&23.2944   &25.5762&22.4985 &24.6459&24.8384&\underline{27.9240}&\textbf{28.4587}\cr
	& SSIM  &0.7098 &	0.7262 &0.8067 &0.6210  &0.8190 &0.7444 &\underline{0.8817} &\textbf{0.8910 }\cr
    \hline

    \multirow{3}{*}{80\%}
    & RSE   &0.1405  &	0.1548 &0.1179 &0.1481  &0.1150	&0.1243 &\underline{0.0893}&\textbf{0.0846 }\cr
	& PSNR	&21.7590 &	20.8928&23.5539&21.2186	&23.4593&22.8495&\underline{25.9261}&\textbf{26.4070}\cr
	& SSIM &0.5895   &	0.5981 &0.7491 &0.5366  &0.7377 &0.6498 &\underline{0.8304} &\textbf{0.8391 }\cr
    \hline

    \multirow{3}{*}{90\%}
    &RSE  	&0.1927 &0.2261 &0.2119 &	0.2856 &0.1541 &0.1654 &\underline{0.1203} &\textbf{0.1163 }\cr
	& PSNR  &18.9587&17.6125&18.0841&15.6648   &21.0432&20.3709&\underline{23.1730}&\textbf{23.4691}\cr
	& SSIM  &0.4127 &0.4115 &0.5453 & 	0.2713 &0.6124 &0.4993 &\underline{0.6807} &\textbf{0.7399 }\cr
    \bottomrule
    \end{tabular}
    \end{threeparttable}
\end{table*}

Fig. \ref{color} displays the visual recovery results on the Lena and Baboon images with 90 percent elements missing.
The images recovered by FBCP, SiLRTC, and TMac are so blurred that we can barely observe the details. For the STDC, there is noticeable accumulated noise, which largely degrades the image quality. Though TMac-TT can recover images decently, severe blocking artifacts largely degrade the visual effect. {In contrast, the proposed TMac-TT+OKA and TWMac-TT+OKA prevail against all the other methods and remove the blocking artifacts.} TMac-TT+KA performs the second best thanks to its higher order and strong low-rankness in balanced modes. While TMac-TT+RE also has a higher-order input, its tensor augmentation method does not have physical meaning. In other words, the reshaping cannot utilize the correlation of different qubits~\cite{latorre2005image}, which corresponds to the TT rank. Therefore, {it is better to input a high-order tensor augmented by KA into the same state-of-the-art TMac-TT algorithm compared to the reshape.}

Meanwhile, by comparing the visual effects in Fig. \ref{color} (f), (g), and (h), we can conclude that the proposed augmentation method OKA performs best when it comes to the tensor augmentation pre-processing methods. OKA overcomes the visual flaws caused by {reshaping} and eliminates the blocking artifacts introduced by KA. Finally, integrated with the OKA scheme, our element weight algorithm TWMac-TT+OKA further improves image recovery. For example, the first row in Fig. \ref{color} exhibits the completion results for classic Lena by different algorithms. If we zoom in the first row {or look at the locally zoomed-in figures in the second row}, there is an apparent difference between TMac-TT+OKA and TWMac-TT+OKA. In addition to the overlapping idea's {effectiveness}, the element-wise weighting scheme further suppresses the local noise and results in a more realistic recovery.

Table {\ref{tab:rgb}} presents the average quantitative results for the five image data. {We bold the optimal values and underline the suboptimal values.} In all the cases, our TWMac-TT+OKA algorithm outperforms all the other compared algorithms in terms of all the evaluation measures, which is consistent with the conclusion reached from the visual results in Fig. \ref{color}. The {second best} result is achieved by TMac-TT+OKA, which is significantly superior to other algorithms. We also observe that the superiority of TWMac-TT+OKA over TMac-TT+OKA concerning the quantitative metrics is relatively slight. 
{This observation demonstrates that the OKA scheme influences more in our final model.} {However, we find that the visual effect (see Fig. \ref{color}) can be improved by using the element weights, so can the quantitative evaluation (as shown in Table \ref{tab:rgb}).} {On the other hand, the weight assignment scheme has a noticeable improvement compared with other baseline models.} The effectiveness of the element-wise weighting idea has also been demonstrated in the synthetic data completion in {Section} \ref{sec_syn}.

Taking the recovery of the Lena image with a 90 percent missing rate as an example, TWMac-TT+OKA achieves the best result among the algorithms, with $RSE \approx 0.0852$, $PSNR \approx 26.5126$ and $SSIM \approx 0.8090$. Comparatively, the result  obtained by the baseline TMac-TT+KA is $RSE \approx 0.1094$, $PSNR \approx 24.3365$ and $SSIM \approx 0.7064$. We obtain an approximately 22 percent improvement over the best current algorithm in terms of the RSE, a 9 percent increase in the PSNR, and an 15 percent increase in the SSIM. When it comes to the average gain obtained by TWMac-TT+OKA over the baseline TMac-TT+KA on all the evaluated images under the 90 percent missing rate situation, we acquire 25 percent gain in the RSE, 12 percent increase in the PSNR and 21 percent in the SSIM. The huge improvement in evaluation indicators proves the superiority of our algorithm on real-world RGB images.

\subsection{Face images under varying illuminations}

\begin{figure*}[htb]
\centering
    \includegraphics[width=0.7\linewidth]{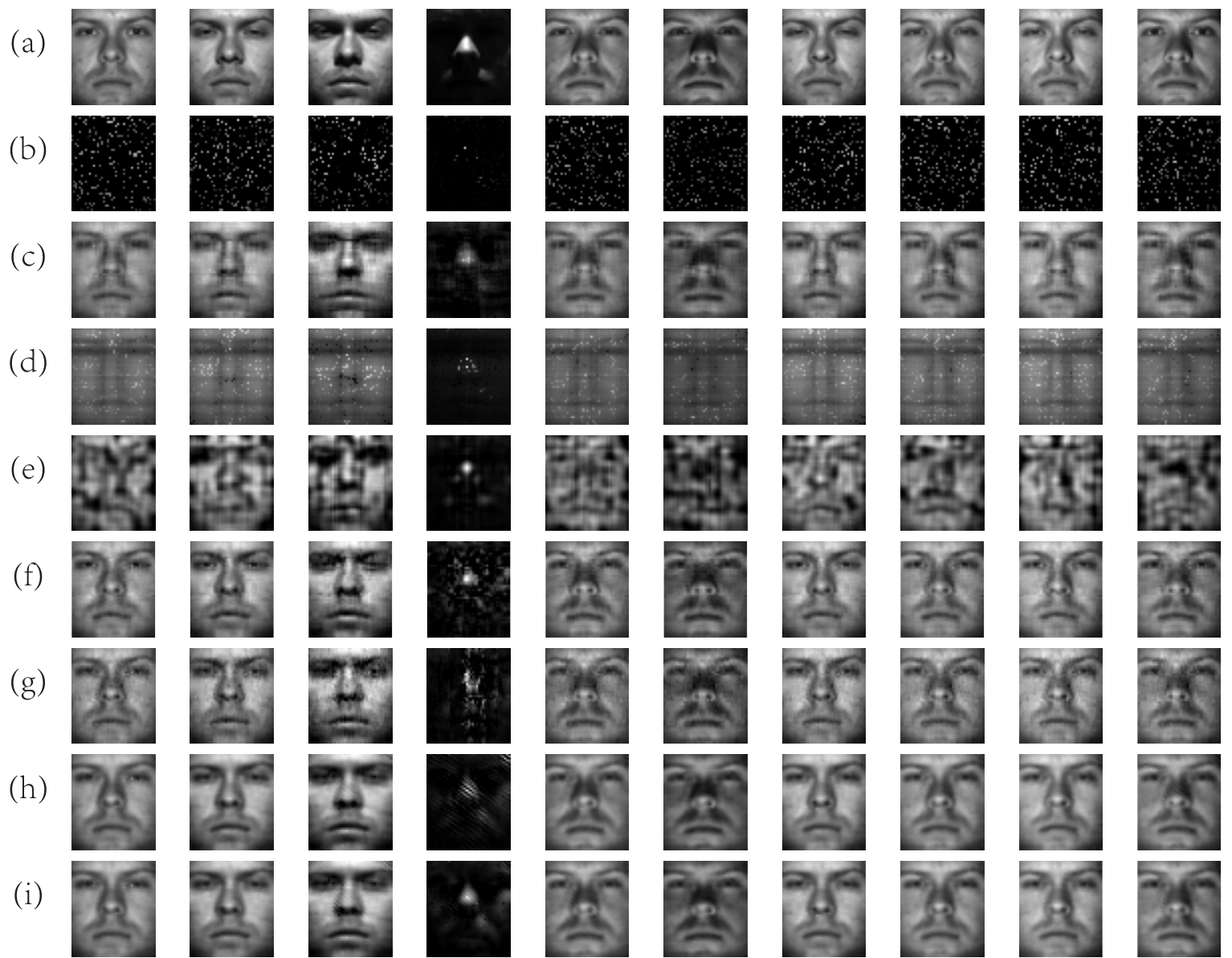}
\caption{Recovery of the Extended Yale B faces with 90\% of missing elements using different algorithms. (a) Original, (b) missing, (c) FBCP, (d) SiLRTC, (e) STDC, (f) TMac, (g) TMac-TT+RE, (h) TMac-TT+OKA, and (i) TWMac-TT+OKA. }
\label{yale90}
\end{figure*}

We test the algorithms on the Extended YaleFace Dataset B, which includes 38 people with nine poses under 64 illumination conditions. This data set is different from RGB images as the channels change from three colors to multiple illuminations. To reduce the computations, we down-sample the original $192\times 162$ images into cropped images of size $48\times 42$. Furthermore, only the frontal pose is used for the test. Thus the input tensor is of size $48\times 42 \times 64$. In this case, KA failed to increase the order of tensors as it is designed only for tensors of size $2^n\times 2^n \times m$, where $n, m \in \mathbb{N}^{+}$ are positive integers and {$(n+1)$} is the target higher order. Thus, we only compare the reshaping and OKA. Reshaping gives a $6\times 8\times 6\times 7 \times 64$ sixth-order tensor, and OKA outputs an eighth-order tensor of size $4\times 4\times 4\times4\times4\times4\times4\times64$.

In Fig. \ref{yale90}, the performance of the algorithms on the face image completion task is shown. The SiLRTC and STDC can barely recover the corrupt faces. By contrast, FBCP and TMac perform much better than SiLRTC and STDC, but the imputation of the missing entries is not accurate at all. Although TMac-TT+RE utilizes the power of both the higher-order and effective tensor train ranks, it does not improve more compared to the classic algorithms. Instead, as long as we replace the reshaping with the proposed OKA technique, namely, TMac-TT+OKA, the recovery quality can be significantly improved, which again demonstrates the effectiveness of OKA.
On the other hand, incorporating the technique of element-wise weighting further enhances overall performance. For example, the image recovered by TMac-TT+OKA in the fourth column has an apparent { striped noise} due to the inflexible mode weights. The proposed TWMac-TT+OKA fixes this phenomenon. The reflection on the nose is distinct, and the shadow on the eyes and cheeks is also evident. Generally, the Yale faces under all the different illuminations are well recovered by the proposed algorithm. TWMac-TT+OKA achieves the best visual result among all the algorithms.

\renewcommand{\arraystretch}{1.2} 
\begin{table}[htp]
 \caption{The averaged recovery performance (RSE, PSNR, SSIM) on the selected yale face under varying illuminations
with missing ratio of 90 percent.}
\label{tab_yale}
  \centering
 \fontsize{7}{8}\selectfont
  \begin{threeparttable}
  \label{Yale_performance_comparison}
    \begin{tabular}{p{2.5cm}p{1cm}p{1cm}p{1cm}}
    \toprule
     &RSE&PSNR&SSIM
    \cr
    \midrule
    FBCP&0.1696 &23.7424 &0.6800 \cr
	SiLRTC&0.3624 &17.1486 &0.4856 \cr
	STDC&0.3462 &17.5440 &0.6178 \cr
	TMac &0.1815 &23.1565 &0.6804 \cr
	TMac-TT+RE &0.1830 &23.0809 &0.7146 \cr
	TMac-TT+OKA&\underline{0.1381} &\underline{25.5300} &\underline{0.7806} \cr
	TWMac-TT+OKA  & \textbf{0.1333}& \textbf{25.8320}&\textbf{0.7908} \cr
    \bottomrule
    \end{tabular}
    \end{threeparttable}
\end{table}

The quantitative results in {Table \ref{tab_yale}} show that the proposed method performs the best in terms of RSE, PSNR and SSIM. The results demonstrate the superiority of the element-wise method in modeling the errors and the weights of recovered elements as TWMac-TT+OKA outperforms TMac-TT+OKA both visually and quantitatively. Comparing the three TT-based algorithms, namely TMac-TT+RE, TMac-TT+OKA, and TWMac-TT+OKA, we find that replacing the reshaping with OKA improves all the quantitative performance to a large extent. Introducing the weighting strategy further improves these three evaluation metrics. Therefore, we arrive at the conclusion that integrating the element-wise weighting and OKA makes a significant contribution to the TT-based algorithm.

\subsection{MRI data completion}\label{mri_sec}
\begin{figure}[htp]
  \centering
  \includegraphics[width=3in]
  {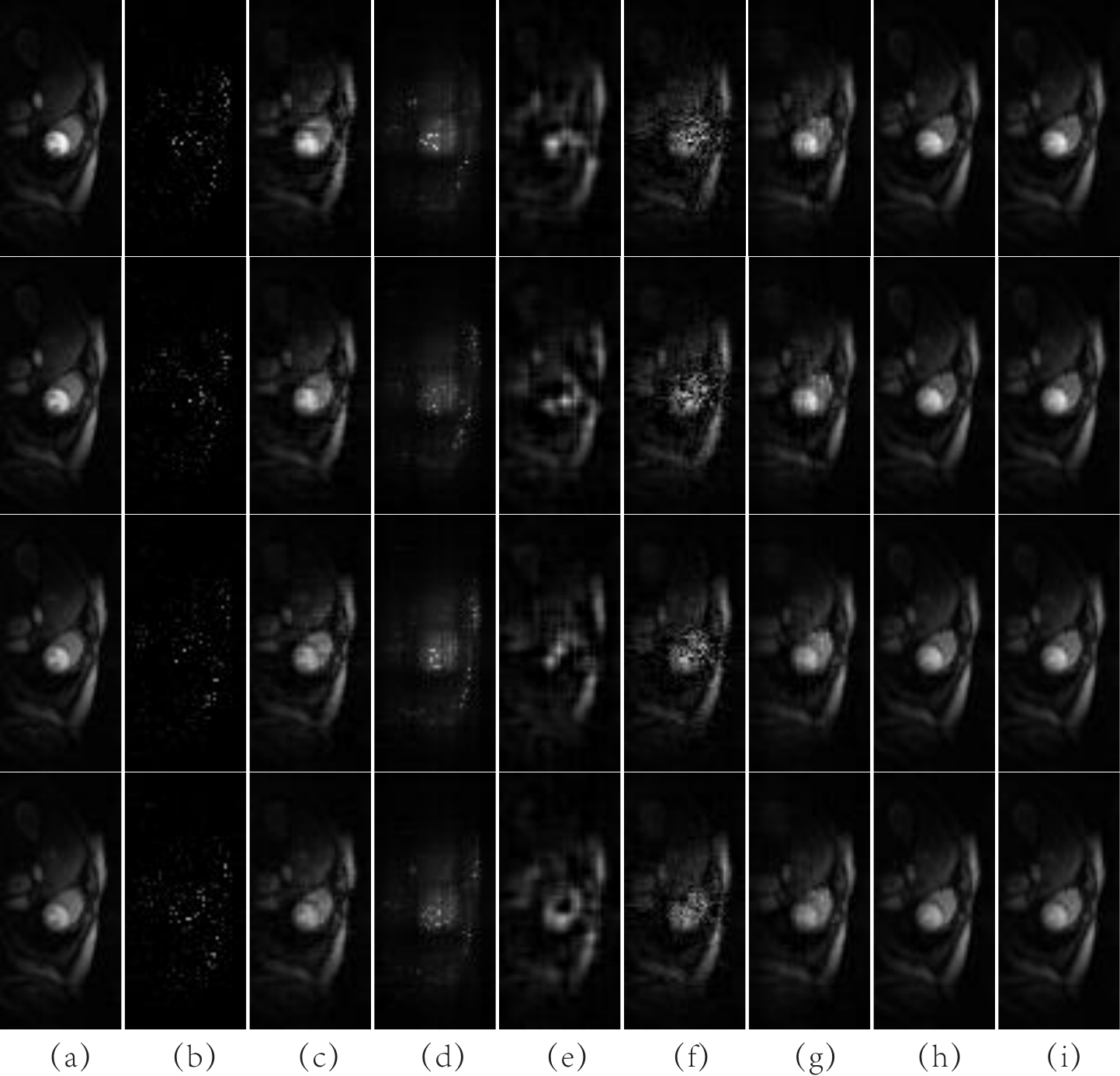}
  \caption{Completion of MRI with 90\% missing elements. The figures from from left to right are:  (a) Original, (b) Missing, (c) FBCP, (d) SiLRTC, (e) STDC, (f) TMac, (g) TMac-TT+RE, (h) TMac-TT+OKA, and (i) our TWMac-TT+OKA. From up to bottom, the shown figures are from the 1st to 4th slices from the selected 25 slices.}
  \label{MRI}
\end{figure}
Magnetic resonance imaging (MRI) data is a natural third-order tensor where the first two indices are for spatial variables and the third index is for object slices. We choose a $64\times 64\times 25$ three-dimensional brain MRI tensor for comparison. In this case, reshaping turns the tensor into a $4\times 4 \times 4 \times 4\times 4 \times 4 \times 25$ seventh-order tensor, and OKA makes the tensor into an eighth-order tensor of size $4\times 4 \times 4 \times 4\times 4 \times 4 \times 4 \times 4\times25$. We show the qualitative results in Fig. \ref{MRI} and the quantitative results in {Table \ref{tab_mri}}.

\renewcommand{\arraystretch}{1.2}
\begin{table}[htp]
 \caption{The averaged recovery performance (RSE, PSNR, SSIM) on the MRI data
with missing ratio of 90 percent.}
\label{tab_mri}
  \centering
 \fontsize{7}{8}\selectfont
  \begin{threeparttable}{}
  \label{MRI_performance_comparison}
    \begin{tabular}{p{2.5cm}p{1cm}p{1cm}p{1cm}}
    \toprule
     &RSE&PSNR&SSIM
    \cr
    \midrule
    FBCP       &0.2461&30.5639&0.8756\cr   
    SiLRTC     &0.5487&23.5973&0.6483 \cr 
	STDC       &0.4632&25.0689&0.7358\cr 
	TMac       &0.4595 &25.1388& 0.7700\cr 
	TMac-TT+RE &0.2134 &31.7999&0.8950\cr  
  TMac-TT+OKA &\underline{0.1373}&\underline{35.6295}& \underline{0.9564}\cr
    TWMac-TT+OKA&\textbf{0.1308}&\textbf{36.0551}&\textbf{0.9576}  \cr

    \bottomrule
    \end{tabular}
    \end{threeparttable}

\end{table}

From Fig. \ref{MRI}, generally, FBCP, TMac-TT+RE, TMac-TT+OKA, and TWMac-TT+OKA show relatively good performance. When looking at the details, we observe that in the first slice (first row in Fig.~\ref{MRI}), only the proposed method TMac-TT+OKA and TWMac-TT+OKA recover the original image without any blur or distortion. The numerical evaluation in {Table \ref{tab_mri}} is also consistent with the analysis above. The proposed TWMac-TT+OKA outperforms all the algorithms in terms of the RSE, PSNR, and SSIM. And the TMac-TT+OKA performs second-best in all these evaluation indices. The RSE achieved by the proposed TWMac-TT+OKA is 39 percent lower than that achieved by the baseline algorithm TMac-TT+RE, which is quite a prominent promotion. The PSNR and SSIM of the proposed TWMac-TT+OKA are also 13 percent and 7 percent higher than those of TMac-TT+RE, respectively. The difference between the TMac-TT+OKA and TWMac-TT+OKA is not distinguishable by the naked eye in the qualitative results. It is difficult to judge which one is better may attribute to the lower input resolution. Still and all, the quantitative results indicate that our final model TWMac-TT+OKA performs better.

\section{Conclusion}\label{conclusion}

This work proposes a novel model named EWLRTC-TT to deal with the LRTC problem based on TT decomposition. To effectively solve this model, the proposed algorithm named TWMac-TT-OKA uses the weighed multilinear matrix factorization technique. To the best of our knowledge, this is the first work that incorporates weighting procedure into multilinear matrix factorization. The proposed algorithm is applied to both synthetic and real-world data represented by higher-order tensors. Extensive experimental results demonstrate that our algorithm is superior to other competing ones and is also highly scalable to various tensors no matter what size they are. In the future, we {may} incorporate our weighted matrix factorization procedure to enhance the performance of the recently proposed tensor completion based on tensor ring decomposition~\cite{zhao2016tensor}.

\ifCLASSOPTIONcaptionsoff
  \newpage
\fi

\bibliographystyle{IEEEtran}

\bibliography{IEEEabrv,TTshort}

\end{document}